\documentclass[sigconf]{acmart}

\AtBeginDocument{%
  \providecommand\BibTeX{{%
    \normalfont B\kern-0.5em{\scshape i\kern-0.25em b}\kern-0.8em\TeX}}}

\setcopyright{rightsretained}

\newcommand{\xhdr}[1]{\noindent\textbf{#1.}}
\usepackage{graphicx}
\usepackage{xcolor}
\usepackage{balance}

\usepackage{natbib}
\usepackage{subcaption}
\usepackage{multirow}
\usepackage{wrapfig}
\usepackage{url}
\usepackage{dsfont} %

\newtheorem{definition}{Definition}

\newcommand{\reals}{\mathbb{R}}

\newcommand{\vx}{\mathbf{x}}

\newcommand{\vz}{\mathbf{z}}

\newcommand{\hy}{\hat{y}}

\newcommand{\vdelta}{\boldsymbol{\delta}}
\newcommand{\vtheta}{\boldsymbol{\theta}}

\newcommand{\vupsilon}{\boldsymbol{\upsilon}}

\newcommand{\cH}{\mathcal{H}}

\newcommand{\cY}{\mathcal{Y}}
\newcommand{\cX}{\mathcal{X}}

\definecolor{mygreen}{RGB}{0,100,0}
\definecolor{myred}{RGB}{139,0,0}

\usepackage[most]{tcolorbox}

\newcommand{\hhl}[1]{\textcolor{black}{#1}}
\newcommand{\hha}[1]{\textcolor{black}{#1}}

\begin{document}
\lhead{}
\rhead{}
\chead{\normalsize In the 4\textsuperscript{th} AAAI/ACM Conference on Artificial Intelligence, Ethics, and Society (AIES-2021)\hrule}

\copyrightyear{2021}
\acmYear{2021}
\acmConference[AIES '21]{Proceedings of the 2021 AAAI/ACM Conference
  on AI, Ethics, and Society}{May 19--21, 2021}{Virtual Event, USA}
\acmBooktitle{Proceedings of the 2021 AAAI/ACM Conference on AI,
  Ethics, and Society (AIES '21), May 19--21, 2021, Virtual Event,
  USA}\acmDOI{10.1145/3461702.3462583}
\acmISBN{978-1-4503-8473-5/21/05}

\title{
  A Human-in-the-loop Framework to Construct Context-aware Mathematical Notions of Outcome Fairness}

\author{Mohammad Yaghini}
\affiliation{%
  \institution{University of Toronto}
  \city{Toronto}
  \country{Canada}}
\email{mohammad.yaghini@mail.utoronto.ca}

\author{Andreas Krause}
\affiliation{%
  \institution{ETH Z{\"u}rich}
  \city{Z{\"u}rich}
  \country{Switzerland}}
\email{krausea@ethz.ch}

\author{Hoda Heidari}
\affiliation{%
  \institution{Carnegie Mellon University}
  \city{Pittsburgh, PA}
  \country{USA}}
\email{hheidari@cmu.edu}

\renewcommand{\shortauthors}{Yaghini et al.}

\begin{abstract}
Existing mathematical notions of fairness fail to account for the \textbf{context} of decision-making. We argue that moral consideration of contextual factors is an inherently \emph{human} task. So we present a framework to learn \emph{context-aware} mathematical formulations of fairness by eliciting people's \emph{situated fairness assessments}. Our family of fairness notions corresponds to a new interpretation of economic models of \emph{Equality of Opportunity (EOP)}, and it includes most existing notions of fairness as special cases. Our \emph{human-in-the-loop} approach is designed to learn the appropriate parameters of the EOP family by utilizing human responses to pair-wise questions about decision subjects' \emph{circumstance} and \emph{deservingness}, and  the \emph{harm/benefit} imposed on them. We illustrate our framework in a hypothetical criminal risk assessment scenario by conducting a series of human-subject experiments on Amazon Mechanical Turk. Our work takes an important initial step toward empowering stakeholders to have a voice in the formulation of fairness for Machine Learning.
\end{abstract}



\maketitle
\thispagestyle{fancy}

\pagestyle{empty}
\section{Introduction}

Despite the surge of interest in designing and guaranteeing mathematical notions of fairness for Machine Learning (ML) (see, e.g., ~\citep{zafar2017dmt,kleinberg2016inherent,hardt2016equality}), existing formulations fail to take the decision-making \emph{context} into account and consider the social \emph{backdrop} and \emph{ramifications} of automated decisions for their subjects and society. Such critical considerations vary across domains, cultures, and societies, and therefore, it is a futile attempt to look for one-size-fits-all definitions of fairness that are considered acceptable across a wide range of situations. Even if we assume certain facets of a highly complex concept like fairness can be reasonably approximated and described in mathematical terms---which itself is a significant assumption---the precise formulation must be highly sensitive to the context. Relying solely on the intuitions of ML experts and practitioners to capture the relevant nuances is likely inadequate and ineffective. We argue that instead, ML practitioners must have a framework in place that allows them to elicit relevant human\footnote{The term `human' here refers to \emph{stakeholders}, including impacted communities and individuals, domain experts, practitioners,  policymakers, and beyond.} judgment and utilize it to construct mathematical formalizations of fairness \emph{tailored} to the decision-making domain at hand.  Such a framework empowers individuals and stakeholders to have a voice in the formulation of fairness and offers practitioners valuable tools to \emph{audit} ML-based decision-making alternatives for \emph{un}fairness in outcomes.

To further illustrate the importance of the \emph{context} and human judgments, consider a popular notion of fairness, called ``equality of odds''~\citep{hardt2016equality} \footnote{Equality of odds requires false positive and false negative rates to be equal for all socially salient groups.} in a \emph{hypothetical} recidivism prediction context, where predictions about future crimes directly translate into pretrial decisions about defendants. As we discuss in our Equality of Opportunity (EOP) framework (Section~\ref{sec:EOP}), the equality of odds requirements can be interpreted as making several \emph{implicit moral assumptions}. Namely, all defendants who would commit a crime if granted bail should receive the same (high-risk) prediction---which subsequently leads to jail time. Similarly, all defendants who will not reoffend should receive the same (low-risk) prediction, leading to bail. In other words, when we employ equality of odds as the notion of fairness in this context, it implies that we agree with the following \hha{value-laden} statements: 
``The true label is a good indicator of a defendant's \emph{deserved} pretrial outcome, and the predicted label is a good indicator of the \emph{benefit/harm/utility} imposed on them as the result of the decision they receive''. We argue that ML-practitioners should not make such value judgments in a vacuum without carefully examining whether they reflect stakeholders' situated perceptions of desert and utility. Regarding pretrial decisions, for example, a closer inspection may reveal that in addition to re-offense, people favor taking a defendant's age and family responsibilities into account to reach a just decision about them.

In this work, we present a practical framework to \emph{construct} context-aware mathematical formulations of fairness utilizing people's  \emph{situated judgments} about fairness in outcomes. We first offer a new interpretation of the theoretical model proposed by \citet{heidari2019a}---which shows that most existing notions of algorithmic fairness are special cases of economic models of \emph{Equality of Opportunity (EOP)}. This new interpretation allows us to design questionnaires that elicit human judgments regarding such concepts as decision-subjects' \emph{circumstance}, \emph{desert}, and \emph{utility}, and utilize the answers to learn a context-appropriate formulation within the EOP family. \hha{As a proof-of-concept, we illustrate our proposal by running} human-subject experiments on Amazon Mechanical Turk (AMT) for a \emph{hypothetical} criminal risk prediction scenario \hha{synthesized using the COMPAS dataset~\citep{propublica_data}.} 
\hha{
Before we proceed to the overview of our empirical setup and findings, two remarks are in order:
First, we must emphasize that our use of the COMPAS dataset to generate our questions is solely to ground our questionnaires in a real-world use-case of ML in making high-stakes decisions about human beings. For the lack of better alternatives, many Fair-ML studies have used the COMPAS dataset for demonstrations in the past. By following suit, we do \emph{not} mean to imply that our work tackles the numerous issues and considerations involved in assessing the fairness of ML in the highly-charged and complicated domain of criminal justice.
Second, we utilized crowdsourcing to illustrate the feasibility of our human-in-the-loop proposal in a \emph{cost}- and \emph{time-efficient} manner. We do \emph{not} advocate for crowdsourcing (or, more generally, relying on the average opinion of ordinary people) as a reliable and ethically sound approach to settle moral dilemmas. We will present a more thorough discussion of this point in Section~\ref{sec:discussion}.}

With the above caveats in mind regarding their intent and scope, our experiments are conducted as follows:
In order to learn the notion that is most compatible with people's perception of fairness, we ask our participants to respond to a series of three questionnaires.
Each participant is required to answer a total of 55 questions---all concerning a \emph{hypothetical} criminal risk assessment scenario. A typical question in our experiment provides the participant with information about two hypothetical decision subjects (defendants) \hha{whose features are synthesized using instances in the COMPAS data set}. The participant is then prompted to choose
(a) which of the two defendants he/she believes is more \emph{deserving} of receiving a lenient decision?
(b) which of them he/she believes would \emph{benefit} more from their algorithmic decision? 
Using the participant's answer to these questions, we estimate the mathematical formulation of fairness that best captures his/her judgments in the decision-making context at hand. %
In addition to estimating the fairness notions of each participant, we propose and investigate two methods to aggregate individual participants' notions into a single definition of fairness. In that sense, our work takes an initial step toward \hha{incorporating people's collective judgments into the process of formulating fairness for algorithmic decision-making}.
Our small-scale empirical findings show that participants favor taking factors beyond the defendant's true- and predicted labels (namely, race, gender,  age, and criminal history) into account when formulating fairness and determining his/her deserved sentencing outcome and utility. Importantly, these considerations are not reflected in any of the existing mathematical formulations of fairness that have been extensively debated in the criminal justice context~\citep{kleinberg2016inherent}.

\hha{
\xhdr{Summary of contributions} We offer a well-justified framework to learn a formal notion of (outcome) fairness tailored to the specific decision-making context at hand. We incorporate contextual factors by eliciting human responses to questions about decision-subjects' deservingness, circumstance, and harm/benefit. Our contributions include:
\begin{itemize}
    \item Identifying a well-justified family of fairness notions (EOP) includes many common notions as special cases.
    \item Learning a new fairness notion for a specific context (criminal risk assessment) by eliciting human judgment and utilizing the data to optimize the EOP family's parameters. 
    \item Collecting and analyzing a novel dataset on people's judgments regarding the above use-case of ML through AMT.
    \item Demonstrating that prior notions of fairness fail to adequately capture the participants' situated fairness judgments.
\end{itemize}
}
In conclusion, our work draws attention to two significant gaps in the growing literature on fairness for ML. First, fairness is a highly context-dependent concept, and designing an ethically acceptable formulation of fairness requires carefully examining the domain and accounting for the social- and individual-level implications of automated decisions. Second, accounting for the social context of decisions is an inherently human task, and the machine on its own lacks the faculties to understand and process the social and ethical aspects of its decisions. Therefore, we need a systematic approach to utilize human-judgment. \hha{Our work presents a practical---yet by nature simplified---toolkit to construct fairness formulations that account for the context and background of decisions---a crucial consideration that existing mathematical formulations of fairness entirely ignore. While it will take many additional iterations and refinements to render this framework ready to employ in real-world applications, we believe it offers a concrete step toward empowering people's voice in value-sensitive ML design, and we hope that the variety of critical questions it brings to the fore lead to impactful directions for future work in Fair-ML.} 

\subsection{Related Work}\label{sec:related}

Numerous mathematical definitions of fairness have been recently proposed and studied; examples include demographic parity~\citep{dwork2012fairness}, disparate impact~\citep{zafar2017dmt}, equality of odds~\citep{hardt2016equality}, and calibration~\citep{kleinberg2016inherent}. While each of these notions may seem appealing on their own, they are incompatible with one another and cannot hold simultaneously (see~\citep{kleinberg2016inherent,chouldechova2017fair}). Furthermore, at least in the context of criminal risk assessment, it is far from settled which one of these notions (if any) is the most appropriate measure of algorithmic fairness (see \citep{propublica} and \citep{dieterich2016compas}). 

Several recent papers empirically investigate AI ethics and algorithmic fairness utilizing human-subject experiments. 
MIT's moral machine~\citep{awad2018moral} provides a crowd-sourcing platform for aggregating human opinions on how self-driving cars should make moral decisions by eliciting their preferences through a series of pairwise preference ordering questions. For the same setting, \citet{noothigattu2018voting} propose learning a random utility model of individual preferences, then efficiently aggregating those individual preferences through a social choice function.  
\citet{lee2019webuildai} proposes a similar approach for ethical decision-making in the context of food distribution. Like the above papers, we obtain input from human-participants by asking them \emph{pairwise} questions (i.e., comparing two alternatives from a moral standpoint). The intent and design of our experiments, however, are entirely different from prior work.

On the topic of algorithmic fairness, \hha{\citet{grgic2018human} study the criteria along which people perceive the use of certain features as unfair in making recidivism predictions about defendants (e.g., whether using the feature ``family's criminal background'' violates defendants' privacy.)}
\citet{veale2018fairness} interview \emph{public sector machine learning practitioners} regarding the challenges of incorporating public values into their work.
\citet{holstein2018improving} investigate \emph{commercial product team}s' needs in developing fairer ML through semi-structured interviews. \citet{woodruff2018qualitative} focus on how affected communities feel about algorithmic fairness. Unlike our work, in which the primary focus is on \emph{quantifying} people's perceptions of fairness, the above papers provide \emph{qualitative} suggestions and best-practices.

Several recent articles investigate people's attitudes towards \emph{existing} mathematical formulations of fairness.
\citet{saxena2019fairness} investigate people's attitude toward three notions of \emph{individual fairness} in the context of \emph{loan decisions}. %
 \citet{srivastava2019mathematical} investigate which of the \emph{existing} group-level formulations of fairness best captures people's perception of fairness. %
\citet{saha2020human} investigate non-experts' comprehension of demographic parity and the impact of the application scenario on it.
 To our knowledge, no prior work has attempted to \emph{construct} \emph{context-aware} notions of fairness utilizing human perception of fairness for ML.

\section{An Equality-of-Opportunity Framework} \label{sec:framework}
In this section, we formally define our learning setting and cast the problem of formulating fairness in \emph{outcomes} as estimating the parameters of an EOP-based criterion. \hha{We must emphasize at the outset that our goal is to narrowly capture \emph{distributive} aspects of (un)fairness through algorithms---namely, we are concerned with disparities in \emph{outcomes} and how harms/benefits are allocated to decision-subjects as the result of the ML-based decisions they receive. Our family of fairness notions solely addresses \emph{outcome/distributive} disparities and remains agnostic to the \emph{causes} and \emph{processes} by which such disparities come to be. Such \emph{procedural} considerations, while extremely important, are outside the scope of the current work.}

\xhdr{The learning environment}
Throughout, we consider the following \emph{simplified} automated decision-making setting: a predictive model (e.g., a hypothetical recidivism prediction model) is trained using historical data records (e.g., the attributes and labels of past defendants) through the standard supervised learning pipeline. The predictions made by this model then directly translate into important decisions for never-before-seen subjects (e.g., a future defendant is sent to jail if the model predicts that he/she would re-offend in the future). 
More precisely, let $T=\{(\vx_i,y_i)\}_{i=1}^n$ denote a training data set consisting of $n$ instances, where $\vx_i \in \cX$ specifies the feature vector for individual $i$ and $y_i \in \cY$, the true label for him/her. Unless otherwise specified, we focus on the binary classification task (i.e., we assume $\cY = \{0,1\}$) and assume without of loss of generality that $\cX = \reals^k$. %
Individuals in $T$ are assumed to be sampled i.i.d. from a population. A learning algorithm uses the training data $T$ to fit a predictive \emph{model} (or hypothesis) $h: \cX \rightarrow \cY$ that predicts the label for out-of-sample instances. More precisely, let $\cH$ be the hypothesis class consisting of all the models the learning algorithm can choose from. The learning algorithm receives $T$ as the input and selects a model $h \in \cH$ that minimizes some notion of empirical loss on $T$. For the ease of notation, when the predictive model in reference is clear from the context, we denote the predicted label for an individual with feature vector $\vx$ by $\hy$ where $\hy = h(\vx)$. %

\subsection{Distributive Fairness as Equality of Opportunity}\label{sec:EOP}
Next, we present the Equality-of-Opportunity (EOP) framework for capturing (un)fairness in the outcomes produced by predictive models for decision-subjects. 
The core idea at the heart of EOP is to distinguish between factors that can morally justify inequality in outcomes among decision subjects and factors that are morally irrelevant and, ideally, should \emph{not} impact outcomes. An equal opportunity policy is one through which an individual's outcome only depends on the former---and not the latter---types of factors. Several models have been proposed to translate EOP into precise mathematical terms (see, e.g., ~\citep{lefranc2009equality}). At a high-level, these models break down an individual's attributes into two categories: 
\emph{circumstance} and \emph{desert}\footnote{Existing formulations of EOP use the term ``effort'' or ``effort-based utility'' instead of ``desert'', but we argue that this model fundamental should summarize all factors that are viewed as legitimate sources of inequality in outcomes---even if those factors have nothing to do with the person's literal effort. We propose the use of the concept and terminology of ``desert'' instead to capture our broader interpretation.}, where \emph{circumstance} captures \emph{all} factors that should not affect the individual's outcome; \emph{outcome} refers the harm/benefit or \emph{utility} he/she receives due to the decision made about him/her; and \emph{desert/deservingness} encapsulates \emph{all} factors that can justify unequal utilities among individual decision subjects.   
More precisely, let $c$ denote the \emph{circumstance}, capturing factors that are not considered legitimate sources of outcome inequality (e.g., race in recidivism prediction). Let $d$ denote a \emph{scalar} summarizing factors that are viewed as legitimate sources of inequality (e.g., prior criminal convictions). %
Let $\phi$ denote the policy that governs the distribution of \emph{utility}, $u$, among subjects. An individual subject's utility is a consequence of his/her desert and circumstance and the implemented policy. Formally, let $F^\phi(. \vert c,d)$ specify the cumulative distribution of utility under policy $\phi$ among people with desert level $d$ and circumstance $c$. 
EOP requires that for individuals with similar desert $d$, the distribution of utility should be the same---regardless of their circumstances:
\begin{definition}[Equality of Opportunity (EOP)]\label{def:EOP}
A policy $\phi$ satisfies EOP if for all circumstances $c, c'$ and all desert levels $d$, $F^\phi(. \vert c,d) = F^\phi(. \vert c',d).$
\end{definition}
With similar reasoning as \citet{heidari2019a}'s, we can show that existing notions of fairness can be cast as special cases of our EOP interpretation. (For instance, by assuming a subject's true label $y$ captures their desert, their sensitive group membership is their circumstance, and their predicted label $\hy$ reflects their utility, we obtain equality of odds.)
\subsection{Parameter Estimation for EOP}\label{sec:estimation}
We restrict our attention to the family of fairness models that can be cast as instances of EOP, and set out to estimate the parameters $c$, $d$, and $u$ that best captures the human perception and judgment of fairness---given a particular decision-making context. %
To formulate fairness as EOP, we need to specify which attributes belong to a subject's circumstance ($c$), what constitutes their desert (or deserved outcome) ($d$), and how the decision they receives impacts their utility ($u$). Providing substantive answers to these  questions requires human feedback, so we provide a framework that turns the input from a human participant (indexed by $p$) into a mathematical formulation of fairness. 

\xhdr{Identifying circumstance $c$}
\looseness -1 From each participant $p$, we ask a series of $k$ questions inquiring whether $p$ believes feature $j \in \{1, \cdots, k\}$ can justify inequality in outcomes. Let $\vz_p$ denote the subset of features that $p$ deems morally irrelevant. We treat each possible value that features in $\vz_p$ can take as a distinct circumstance $c$.

\xhdr{Estimating desert $d$}
Consider a participant $p$ and a decision subject with characteristics $\vx$ and $y$. Let $d_p$ specify the subject's \emph{overall desert} according to $p$. We assume $d_p$ is not directly observable, but there exists a function $\delta_p: \cX \times \cY \rightarrow \reals^+$, such that $d = \delta_p(\vx, y).$
That is, $\delta_p$ maps the information the participant observes about the decision subject (i.e., $\vx$ and $y$) to his/her desert. 
For simplicity, we assume $\delta_p$ is a linear function of $\vx, y$, that is, there exists a coefficient vector $\vdelta_p$ such that $d = \vdelta_p. [ \vx, y ]$ (where $[\vx, y]$ is the vector concatenation of feature vector $\vx$ and true label $y$). 
To estimate the $\vdelta_p$, we ask $p$ a series of $Q$ pairwise comparison questions. In each question $q=1,\cdots,Q$, we present $p$ with the information about two hypothetical decision subjects, $i^q_1=[\vx^q_1, y^q_1]$ and $i^q_2= [ \vx^q_2, y^q_2 ]$, and ask her which one of them she considers to be more deserving of receiving the positive prediction.\footnote{Note that different participants may respond to different sets of questions, hence to be precise, all the notation defined in this part must have a $p$ superscript. For brevity, when the participant in reference is clear from the context, we drop the superscript $p$.} We assume $p$ would pick $i^q_1$ with probability 
\begin{equation}\label{eq:normal}
\Phi (\vdelta_p. [ \vx^q_1-\vx^q_2, y^q_1-y^q_2 ])
\end{equation}
where $\vdelta_p$ is the unknown parameter we wish to estimate and $\Phi$ represents the cumulative distribution function of the standard normal distribution. Note that our modeling choices here closely matches prior work (see, e.g., ~\citep{noothigattu2018voting}). 

With this model in place, we can readily find the Maximum Likelihood Estimator (MLE) of $\vdelta_p$ using $p$'s answers to the pairwise questions presented to her. More precisely, let $a^{q}$ be $p$'s answer to the $q$'th pairwise question, in which she has expressed confidence level $c \in \{1, 2\}$. $a^{q}$ is $|c|$ if $p$ chooses subject 1 in response to question $q$, and it is $-|c|$ otherwise. We find the maximum likelihood estimator or $\vdelta_p$ by solving the following optimization:
{\scriptsize
\begin{equation}
\arg \min_{\vdelta}   -\sum_{q=1}^Q \log\Phi \left(a^q\,\vdelta. [ \vx^q_1-\vx^q_2, y^q_1-y^q_2 ] \right)
\text{s. t. } \Vert \vdelta \Vert_2 \leq 1 
\end{equation}
}

\xhdr{Estimating individual utility $u$}
Let $u_p$ denote the \emph{utility} a decision subject $[ \vx,y ]$ earns as a result of being subject to decision rule $h$ according to participant $p$.
For simplicity, we assume $u_p$ is a linear function of $\vx,y,\hy$. That is, there exists $\vupsilon_p$, such that $u_p = \vupsilon_p. [ \vx, y, \hy ]$. 
To estimate $\vupsilon_p$, we ask $p$ a series of $Q$ pairwise comparison questions. In each question $q=1, \cdots, Q$, we present $p$ with the information about two hypothetical decision subjects, $i^q_1=[ \vx^q_1, y^q_1, \hy^q_1 ]$ and $i^q_2= [ \vx^q_2, y^q_2, \hy^q_2 ]$, and ask her which one of them she believes would benefit more from his/her decision. We assume $p$ would pick $i^q_1$ with probability 
 \begin{equation}\label{participant_utility}
\Phi (\vupsilon_p. [ \vx^q_1-\vx^q_2, y^q_1-y^q_2, \hy^q_1-\hy^q_2 ])
\end{equation}
where $\vupsilon_p$ is the underlying parameters we wish to estimate. With this model in place, we can compute the MLE of $\vupsilon_p$ using $p$'s answers to the pairwise questions presented to her.

\subsection{Methods of Aggregation}\label{sec:aggregation}
For each individual participant $p$, we estimate the circumstance, desert, and utility functions that best capture his/her judgments. We will denote these as $\vz_p, \vdelta_p$, and $\vupsilon_p$, respectively. \hha{Using these estimates for $\vz_p, \vdelta_p$, and $\vupsilon_p$, we can construct an EOP-based notion of fairness (Definition~\ref{def:EOP}) according to participant $p$' judgments.} To aggregate these individual notions into one representing society as s whole, we consider two natural aggregation methods.\footnote{Many alterbative aggregation methods are conceivable. For brevity and as reasonable starting points, we only study these two.}
The first method is inspired by Borda Count:%
\begin{itemize}
\item A feature is part of a subject's circumstance if more than half of  participants (or voters) deem it to be so.
\item {\footnotesize $\vdelta^*(\vx,y) = \frac{1}{N}\sum_{p=1}^N  \vdelta_p . [ \vx,y ] =  \left( \frac{1}{N}\sum_{p=1}^N  \vdelta_p \right) .\;[ \vx,y ]$. }
\item {\scriptsize$\vupsilon^*(\vx,y,\hy) = \frac{1}{N}\sum_{p=1}^N  \vupsilon_p . [\vx,y,\hy ] = \left( \frac{1}{N}\sum_{p=1}^N  \vupsilon_p \right).\:[\vx,y,\hy ]$.}
\end{itemize}
The second method is inspired by the hierarchical Bayesian modeling approach. We assume there exists a $\vtheta$ that captures society's perception of desert. 
A participant $p$ in the society has his own $\vdelta_p$ which is a noisy version of $\vdelta$.
To estimate $\vdelta$ and $\vdelta_p$'s jointly, we solve the convex optimization problem:
{\scriptsize
\begin{eqnarray}\label{eq:hierarchical}
\arg \min_{\vdelta_p,\vtheta}  && -\sum_{p}\sum_{q} \log\Phi \left(a^q\,\vdelta_p. [ \vx^{p,q}_1-\vx^{p,q}_2, y^{p,q}_1-y^{p,q}_2 ] \right)  \\
\text{s. t. } 
&& \Vert \vdelta_p - \vdelta \Vert_2 \leq \lambda \text{ and } \Vert \vtheta_p \Vert_2 \leq 1, \; \Vert \vdelta \Vert_2 \leq 1 \nonumber.
\end{eqnarray}
}
We estimate the individual and aggregate utility coefficients (that is, $\vupsilon_p$'s and $\vupsilon$) similarly.

\section{Study Design}\label{sec:methodology}
We conduct a series of questionnaires on AMT to construct a notion of fairness that best captures participants' perception of fairness in a hypothetical crime prediction context. %

\hha{\xhdr{A note on the use of ML in criminal justice} The use of ML for risk assessment in the criminal justice system have been heavily and validly criticized (see, e.g.,~\citep{propublica,monahan2016risk,stevenson2018assessing}). Many scholars argue that such tools must be entirely banished to give way to sorely-needed structural changes in the criminal justice system (including but not limited to bail reform in the pretrial context). Among noteworthy objections to the COMPAS tool, in particular, is its lack of transparency~\citep{rudin2018age}, efficiency gains~\citep{dressel2018accuracy}, and the numerous biases encoded in the data it is based off of~\citep{grgic2018human} (for example, one source of bias is that the target label ``rearrest within two-years'' is an imperfect proxy for ``future crimes''~\citep{wick2019unlocking}). While we do not condone the use of ML in criminal justice processes (in particular, the COMPAS tool for criminal risk assessment), given the current wide-spread use and impact of such systems, we believe it is paramount for the ongoing debate around ML-fairness in this context to be grounded in stakeholders' values and collective moral judgments.}

\subsection{Scenarios and Contexts}
We first introduce the goals and scope of our study to participants as follows:\footnote{Throughout, text boxes display the content exactly as it was shown to our study participants through the user interface.} If the participant needs further information to understand the task, he/she can access a longer version, which includes several examples illustrating the task. 
Figure~\ref{fig:long_intro} in the Appendix shows a longer version of the introduction to the HIT, which includes several examples.

{\small
\begin{tcolorbox}[breakable, enhanced]
\xhdr{Background and Task Description} Data-driven decision-making algorithms are increasingly employed to automate the process of making important decisions for humans, in areas such as credit lending, medicine, criminal justice, and beyond. \textbf{Organizations} in charge of decision-making can utilize massive datasets of historical records to learn a \textbf{decision-making} rule capable of making accurate \textbf{predictions} about never-before-seen individuals. Such predictions often serve as the basis for consequential \textbf{decisions} for these individuals. (From this point on we will refer to \textbf{individuals} subject to decisions as \textbf{decision subjects}.)

In recent years, several studies have shown that \textbf{automated}/\textbf{algorithmic decisions} made in the above fashion may disparately impact certain groups and individuals. For instance, in the context of credit lending, the decision-making rule may systematically disadvantage loan applicants belonging to a certain racial group and reject their loan applications more frequently. These observations have raised many questions and concerns about the fairness of automated decisions.

The goal of our study is to understand your moral reasoning and perception about what it means for automated decisions to be fair---considering the specifics of the decision-making context. We would like to know your ethical judgment through your answers to the following questions:
\begin{itemize}
\item Which \textbf{attributes} of a decision subject do you consider \textbf{morally acceptable} for the decision-making rule to base its decisions on? 
\item Comparing the attributes of two decision subjects, which one of them do you believe is more \textbf{deserving} of receiving a \textbf{better (more desirable) decision}? 
\item How do you believe automated decisions will \textbf{impact} these subjects? We'd like you to imagine how algorithmic decisions may contribute to the overall \textbf{happiness}, \textbf{satisfaction}, \& \textbf{well-being} of a decision subject.
\end{itemize}
\end{tcolorbox}
}
Throughout the experiment, we focus on a criminal risk prediction context, described below:

{\small \begin{tcolorbox}[breakable, enhanced]
\xhdr{Decision-Making Context} In court-rooms across the United States, data-driven decision-making algorithms are employed to predict the likelihood of future crimes by defendants. These algorithmic predictions are utilized by judges to make sentencing decisions for defendants (e.g., setting the bail amount, or time to be spent in jail). Decision-making algorithms use historical data about past defendants to learn about factors that highly correlate with criminality. For instance, the algorithm may learn from past data that: 1) a defendant with a lengthy criminal history is more likely to reoffend if set free on bail---compared to a first-time offender, or 2) defendants belonging to certain groups (e.g., residents of neighborhoods with high crime rate) are more likely to reoffend if set free. These automated predictions may directly translate into sentencing decisions. For instance, a defendant who is predicted to have a high risk of reoffending may be sentenced to jail, whereas a defendant who is predicted to have a low risk of reoffending may be set free on bail.
\end{tcolorbox}}

\subsection{User Interfaces}
Through our \emph{conversational interface}\footnote{Conversational UIs are known to increase the user's attention and feel more natural to work with. See~\citep{rice2019conversational}.} the participant then responds to three questionnaires. The first questionnaire contains five questions, and the second and third parts, each consists of 25 questions. Once the participant completes the main three questionnaires, they can optionally provide us with their demographic information. (Participants had the option to skip this exit questionnaire).

\xhdr{Part 1: Identifying $\vz$}
To determine which attributes of a decision subject belong to the category of morally irrelevant, we first display the following introductory text:

{\small \begin{tcolorbox}[breakable, enhanced]
In the first part, we would like to understand your moral reasoning about the following: 
Which attributes of a defendant do you consider morally acceptable for the decision-making rule to base its predictions on? 

{ \color{mygreen} \textbf{Example:} one may believe it acceptable for the decision-making rule to take the subject's criminal history into account, but find it unacceptable for his/her parents' criminal history to impact whether he/she is predicted to have a high or low risk of reoffending.}

{ \color{myred}(Note that this example is only meant to illustrate the task. You may have a very different opinion.) }
\end{tcolorbox}}

The participant then responds to a simple questionnaire consisting of 5 questions of the following form:

{\small \begin{tcolorbox}[breakable, enhanced]
To what extent do you agree with the following statement:

It is ethically acceptable for the attribute [...] (which can take one of the following values: [...]) to impact the decision a defendant receives. 
\end{tcolorbox}}

\begin{figure}[h!]
    \centering
    \includegraphics[width=.48\textwidth]{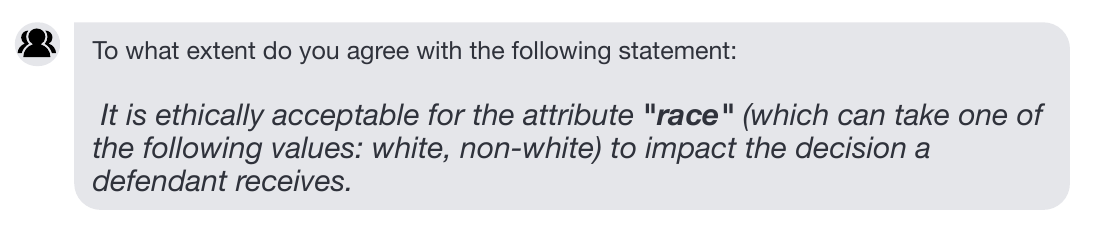}
    \caption{A typical question in part 1 of our experiments.}
    \label{fig:UI_part1}
\end{figure}

For an example of this type of question, see Figure~\ref{fig:UI_part1}. The participant can choose their response from a 4-point Likert scale (``Disagree'', ``Somewhat Disagree'', ``Somewhat Agree'', and ``Agree''). The participant was encouraged (but not required) to provide a justification for their choice in a free-form text. 

\xhdr{Part 2: Estimating $\vdelta$}
In the second part, our goal is to estimate the desert function (i.e., $\vdelta$). We first display the following introductory text:

{\small \begin{tcolorbox}[breakable, enhanced]
In the second part, we would like to understand your moral reasoning about the following: 
Comparing the attributes of two defendants, which one of them do you believe is more deserving of receiving a more lenient decision? 

{ \color{mygreen} \textbf{Example:} Consider two defendants with similar attributes, except for their employment status---one unemployed, the other a local government employee. One may consider the employed subject more deserving of the ``low risk to reoffend'' prediction.}

{ \color{myred}(Note that this example is only meant to illustrate the task. You may have a very different opinion.) }
\end{tcolorbox}}

\begin{figure}[h!]
    \centering
    \includegraphics[width=.48\textwidth]{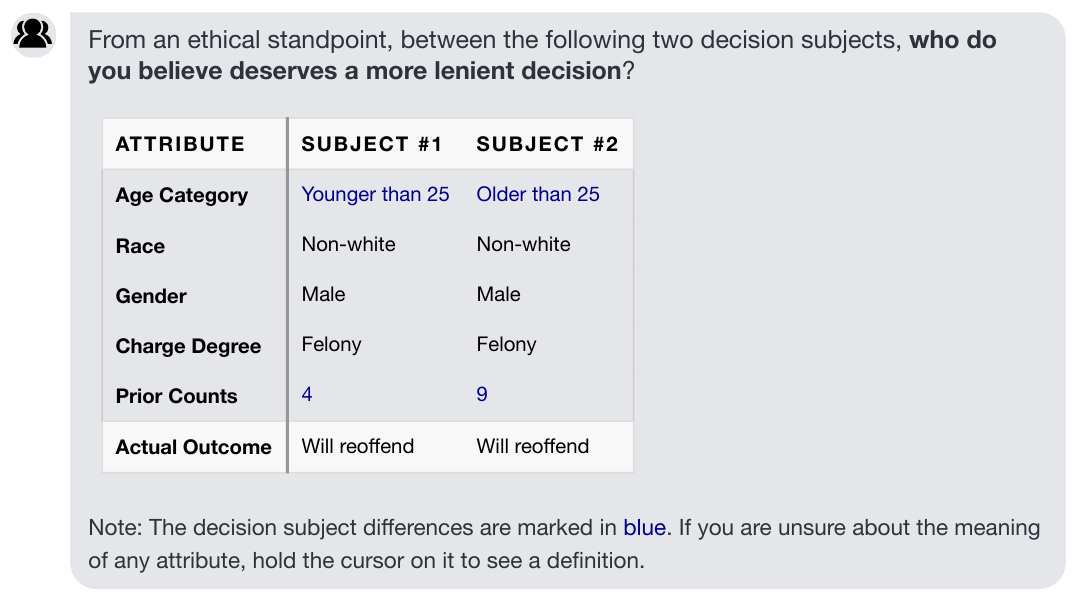}
    \caption{A typical question in part 2 of our experiments.}
    \label{fig:UI_part2}
\end{figure}

We then prompt the participant to respond to a series of pairwise comparison questions of the following form:

{\small \begin{tcolorbox}[breakable, enhanced]
From an ethical standpoint, between the following two decision subjects, who do you believe deserves a more lenient decision? 
\end{tcolorbox}}
For an example of this type of question, see Figure~\ref{fig:UI_part2}. 
The participant can choose either ``Clearly subject 1'', ``Possibly subject 1'', ``Possibly subject 2'', and ``Clearly subject 2''. The participant was encouraged (but not required) to provide a justification for their choice in a free-form text. 

\xhdr{Part 3: Estimating $\vupsilon$}
To estimate the utility function (i.e., $\vupsilon$), we first display the following introductory text:

{\small \begin{tcolorbox}[breakable, enhanced]
In the third part, we would like to understand your moral reasoning about the following: 
Given the attributes of two defendants, which one of them do you believe would benefit more from their respective algorithmic decision? 
In responding to this question, imagine yourself in the circumstances of these two defendants and think about how the sentencing decision they receive may affect their lives.

{ \color{mygreen} \textbf{Example:} Consider two defendants with similar attributes, except for their number of dependants (one with two children and another with no dependents.). One may believe that a ``low risk to reoffend'' prediction would contribute more to the overall satisfaction, happiness, and well-being of the subject who has kids.}

{ \color{myred}
(Note that this example is only meant to illustrate the task. You may have a very different opinion.)}
\end{tcolorbox}}

We then prompt the participant to respond to a series of pairwise comparison questions of the following form:

{\small \begin{tcolorbox}[breakable, enhanced]
From an ethical standpoint, between the two following decision subjects, who do you think will benefit more from their algorithmic decision?
\end{tcolorbox}}

\begin{figure}[h!]
    \centering
    \includegraphics[width=.48\textwidth]{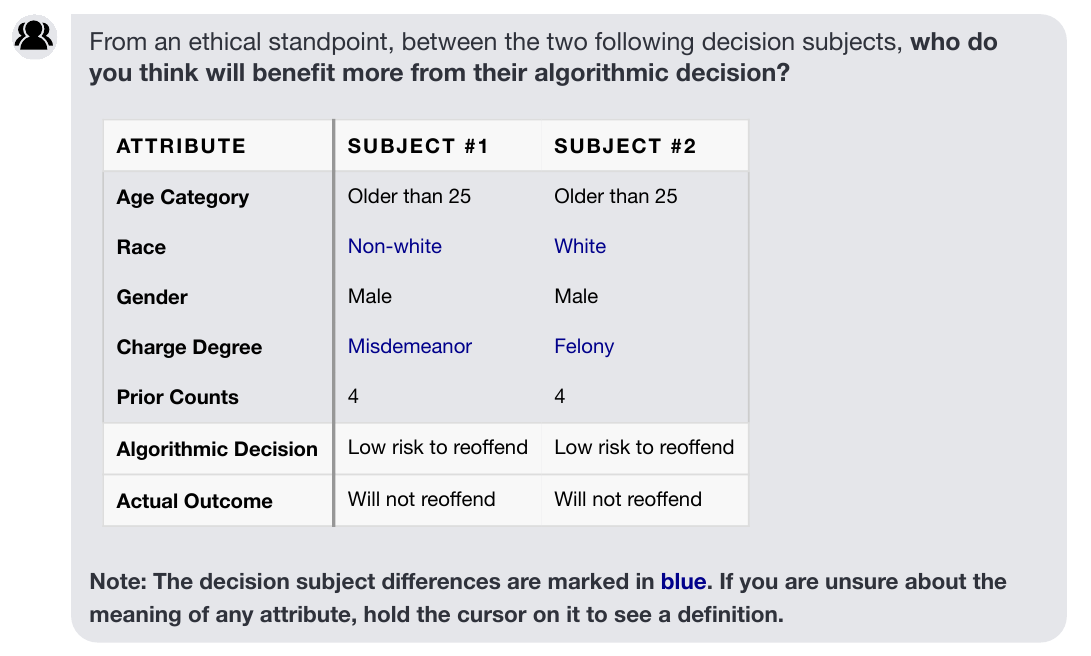}
    \caption{A typical question in part 3 of our experiments.}
    \label{fig:UI_part3}
\end{figure}

For an example of this type of question, see Figure~\ref{fig:UI_part3}. 
The participant can choose either ``Clearly subject 1'', ``Possibly subject 1'', ``Possibly subject 2'', and ``Clearly subject 2''. The participant was encouraged (but not required) to provide a justification for their choice in a free-form text. 

\subsection{Experimental Design}\label{sec:experimental}
We used a summarized and pre-processed version of the COMPAS data set~\citep{propublica} to generate our questions. We restricted attention to the following attributes for each defendant: \textbf{gender} (1 if the defendant is male, 0 otherwise), \textbf{age} (1 if the defendant is less than 25, 0 otherwise), \textbf{race} (1 if the defendant is not Caucasian, 0 otherwise), \textbf{charge degree} (1 if felony, 0 otherwise), \textbf{prior counts} (a positive integer). We set the \textbf{true label} to 1 if the defendant reoffends within 2 years, 0 otherwise). \hha{Note that in the original COMPAS dataset, the true label reflects whether a defendant is \emph{rearrested} for an alleged crime. This is clearly different from our choice of the true label (whether the defendant actually reoffends), which we chose to reduce the layers of complication.} The \textbf{predicted label} indicates if the COMPAS tool gives a score of 5 or higher\footnote{Note that the COMPAS tool bases its risk scores on many features other than the ones considered here. Our focus is not on the COMPAS tool and its inner workings. We simply use the COMPAS risk scores (provided by \citet{propublica}) to generate binary recidivism predictions for defendants in our hypothetical scenario.} to the defendant. 

\begin{figure}
  \begin{minipage}[c]{0.3\textwidth}
    \includegraphics[width=\textwidth]{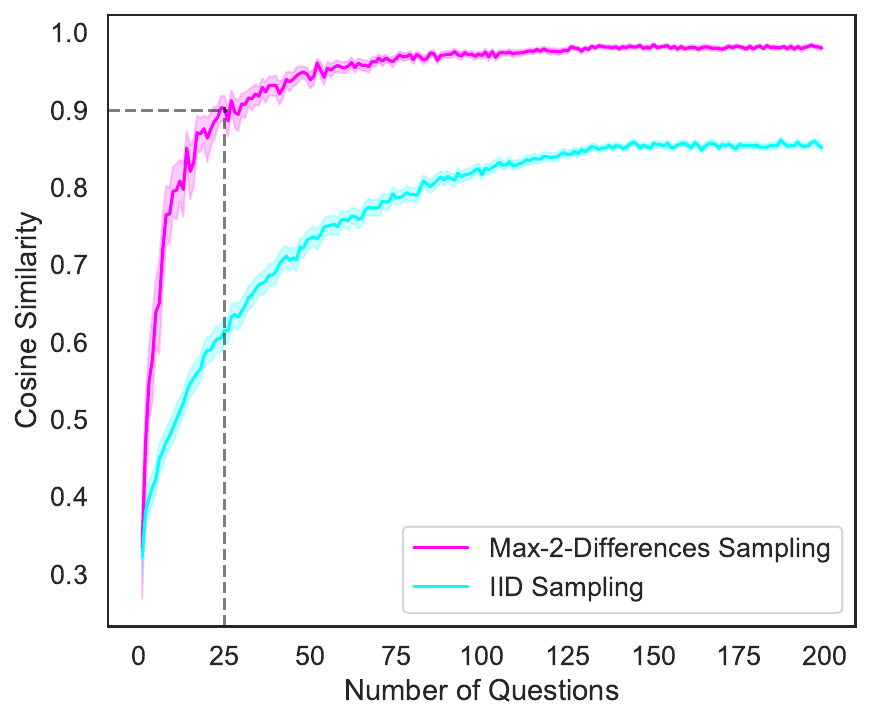}
  \end{minipage}\hfill
  \begin{minipage}[c]{0.15\textwidth}
    \caption{Cosine similarity between the true weight vectors (according to which responses are simulated) and the estimated ones (MLE base on simulated responses).}
    \label{fig:sims}
  \end{minipage}
\end{figure}

In part 1, we asked one question for each of the five features in the data set.
In part 2, to generate a question $q$, we picked the first defendant ($i^q_1$) i.i.d. from the data set, then chose the second defendant ($i^q_2$) such that it does not differ with the first one in more than two attributes. The reason for this restriction was two-fold: (1) As shown in Figure~\ref{fig:sims} it allows us to recover the underlying weight vector confidently with a smaller number of questions; (2) it reduces the cognitive burden of comparing the two defendants with one another for the participants\footnote{Comparing two objects that differ across multiple dimensions is a type of \emph{multitasking}---which notoriously reduces human attention and leads to erroneous responses. See \url{https://en.wikipedia.org/wiki/Human_multitasking} and the references therein.}.
Questions in part 3 were generated similarly. 
Part 2 and 3 each consisted of 25 questions. We chose the number of questions by simulating our model of participants responses (Equation~\ref{eq:normal}) and finding the number of questions necessary to learn their weight vectors with a satisfactory accuracy (i.e., 90\%). See figure~\ref{fig:sims}.

We validated our interface design through two rounds of \emph{pilot} studies---one internal (among members of our research group) and one on AMT (among 20 crowd-workers). Based on the feedback we received from pilot participants, we made several (minor) changes to the text to improve readability and allowed participants to edit their responses. We also decided to randomize the order in which questionnaires in part 2 and 3 are shown to the participants, and included one attention-check question in each part.

\section{Experiments}\label{sec:findings}
To illustrate our framework, we conduct a series of questionnaires on AMT to construct a notion of fairness that best captures people's perceptions in our hypothetical criminal risk prediction context. %
Through our experiments on AMT, we gathered a data set consisting of 90 participants' responses to our questionnaires.
As a final step, we asked the participants to provide us with their demographic information, such as their age, gender, race, education, and political affiliation. The purpose of asking these questions was to understand whether there are systematic variations in perceptions of fairness across different demographic groups.\footnote{Answering to this part was entirely optional and did not affect the participant's eligibility for monetary compensation.}
Table~\ref{tab:demographic} summarizes the demographic information of our participants and contrasts it with the 2016 U.S. census data. In general, AMT workers are not a representative sample of the U.S. population (they have Internet access and are willing to complete online tasks). In particular, our participants were younger and more liberal compared to the average U.S. population. \hha{The majority of our participants were white, with Black and Hispanic minorities underrepresented compared to census numbers. (As we mentioned earlier, our experiment is meant as a proof-of-concept; the proper representation of disadvantaged minorities is one of the essential preconditions for the ethical use of such participatory frameworks in the real world.)}

\begin{table}[h]
\caption{Demographics of our AMT participants compared to the 2016 U.S. census data.}
\label{tab:demographic}
\centering
\begin{tabular}{l l l l}
Demographic Attribute & AMT & Census\\ \hline \hline
Male  & 61\% & 49\% \\
Female  & 39\% & 51\% \\\hline
Caucasian  & 73\% & 61\% \\
African-American  & 9\% &  13\%\\
Asian  & 10\% &  6\%\\
Hispanic  & 4\% &  18\%\\ \hline
Liberal  & 65\% &  33\%\\
Conservative  & 24\% &  29\%\\ \hline
High school  & 34\% &  40\%\\
College degree  & 58\% &  48\%\\
Graduate degree  & 8\% &  11\%\\ \hline
18--25  & 12\% & 10\%\\ 
25--40  & 65\% & 20\% \\ 
40--60  & 17\% &  26\%\\ \hline
\end{tabular}
\end{table}

\begin{figure}
  \begin{minipage}[c]{0.3\textwidth}
    \includegraphics[width=\textwidth]{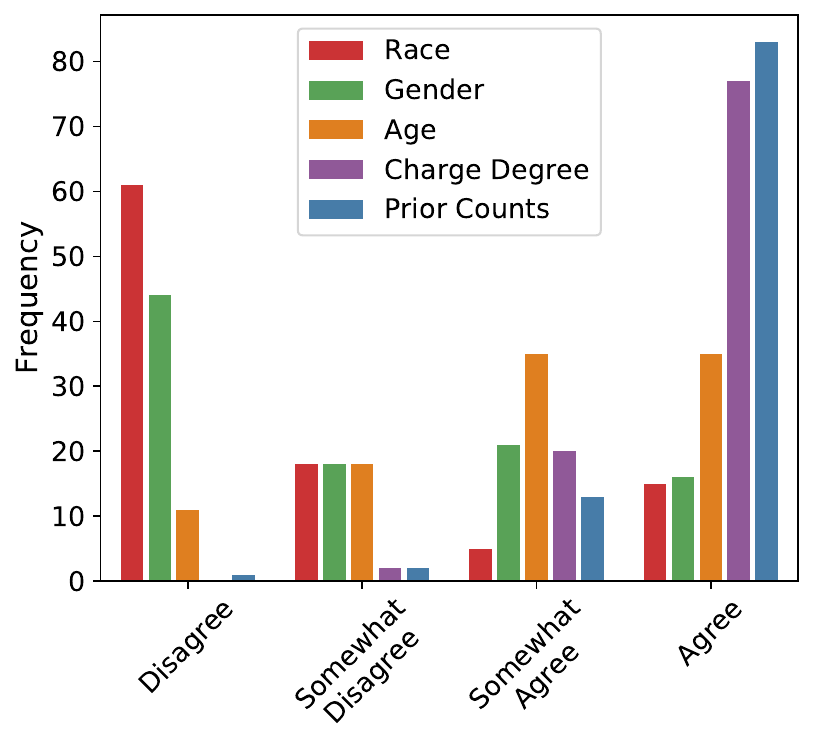}
  \end{minipage}\hfill
  \begin{minipage}[c]{0.15\textwidth}
	\caption{Responses to part 1 questions. Participants believe the number of prior counts, charge degree, \& age can morally justify inequality in sentencing decisions, while race and gender are morally irrelevant.}
	\label{fig:part1}
  \end{minipage}
\end{figure}

\xhdr{Part 1 results: Identifying morally irrelevant features}
Figure~\ref{fig:part1} shows the distribution of responses to questions in part 1 of our experiment. Responses represent participants' extent of agreement with the statement ``It is ethically acceptable for the attribute [...] to impact the decision a defendant receives?'' The majority of participants strongly believe the number of prior counts and charge degree can morally justify inequality in sentencing decisions. The majority also strongly believe that race is morally irrelevant. For gender and age, opinions are mixed. On average, participants consider gender as part of a defendant's circumstance, and age as a morally relevant feature, but their degree of confidence is generally lower compared to the other three features.

\begin{table}[h]
\caption{The average coefficient of each attribute in our estimated desert and utility functions.}
\label{tab:avg_weights}
\centering
{ \scriptsize
\begin{tabular}{lccccccc}
  &  Race &  Gender &  Age & Charge & Prior & $y$ & $\hy$ \\ \hline
$\delta$ &  $0.11$ &  $-0.19$ &  $0.13$ &  $-0.46$ &  $-0.25$ & $-0.41$ & N/A\\
$\upsilon$ & $0.06$ & $-0.07$  &  $0.09$ & $-0.09$  &  $-0.08$ & $-0.18$ & $-0.43$\\
\end{tabular}
}
\end{table}

\begin{figure*}
\centering
\begin{minipage}{.3\textwidth}
  \centering
		\includegraphics[width=\textwidth]{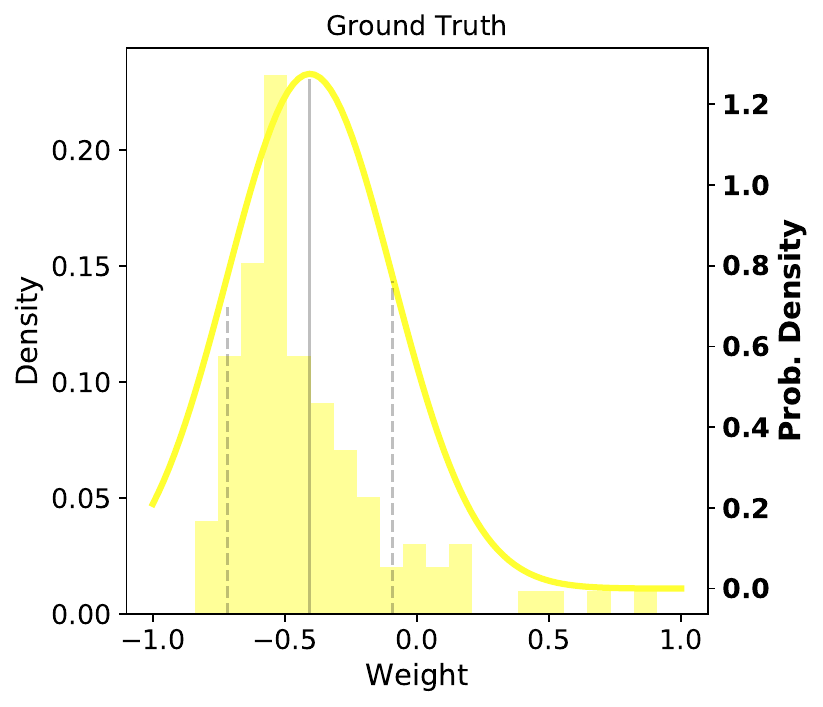}
		\caption{The distribution of $\delta$ for the true label ($y$) when $\vdelta$ is estimated for each participant, separately.}
		\label{fig:y_weights_desert}
\end{minipage}\hspace{2mm}
\begin{minipage}{.3\textwidth}
  \centering
		\includegraphics[width=\textwidth]{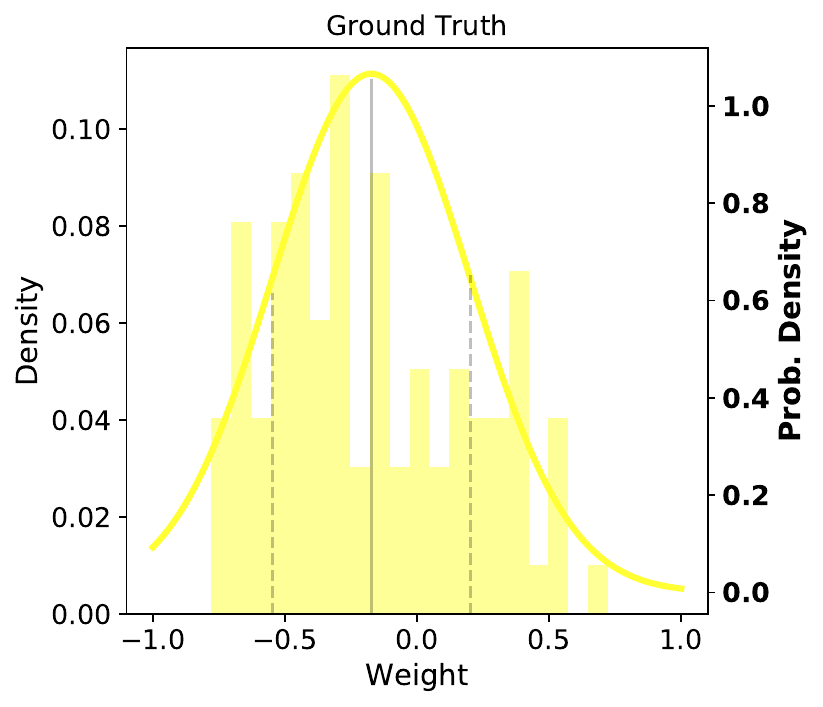}
		\caption{The distribution of $\upsilon$ for the true label ($y$) when $\vupsilon$ is estimated for each participant, separately.}
		\label{fig:y_weights_utility}
\end{minipage}\hspace{2mm}
\begin{minipage}{.3\textwidth}
  \centering
	   \includegraphics[width=\textwidth]{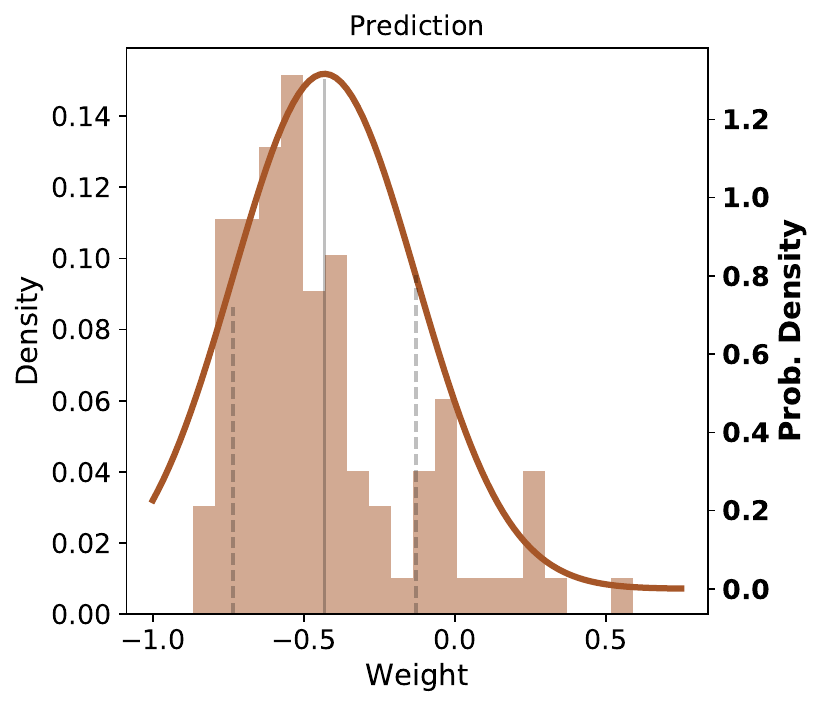}
    	\caption{The distribution of $\upsilon$ for the predicted label ($\hy$) when $\vupsilon$ is estimated for each participant, separately.} 
    	\label{fig:yhat_weights_utility}
\end{minipage}\hspace{2mm}
\end{figure*}

\xhdr{Part 2 results: Estimating the desert function ($\vdelta$)}
Figure~\ref{fig:y_weights_desert} illustrates the distribution of weights allocated to the true outcome $y$ when estimating $\vdelta$. As one may expect, the true outcome has a significant weight, but somewhat surprisingly, on average participants gave similar weights to charge degree (-0.46) and the true label (-.41).
Figure~\ref{fig:feature_weights} (left panel) in Appendix~\ref{app:empirical} shows the distribution of weights given to the five features of a defendant (race, gender, age, charge degree, and prior counts) when estimating desert ($\vdelta$). 
\hhl{
(For ease of interpretation, a Gaussian probability density function with the same mean and variance as the data is plotted on the opposite axis with bolder typeface. We have used 20 bins for these illustrations.) 
}
According to our participants, younger defendants generally deserve a more lenient sentencing decision. The same holds if a defendant is female or non-white. Participants consider a defendant less deserving of a lenient outcome if they are charged with a felony, or they have a large number of prior convictions. Perhaps surprising, participants, on average, gave a larger weight (-0.25) to charge degree%
 as opposed to prior \emph{conviction} counts (-0.46).

\xhdr{Part 3 results: Estimating the utility function ($\vupsilon$)}
Figure~\ref{fig:y_weights_utility} illustrates the distribution of weights allocated to the true outcome $y$ when estimating $\vupsilon$ and Figure~\ref{fig:yhat_weights_utility} shows the distribution of weight on the predicted outcome, $\hy$. The contribution of $\hy$ is significantly larger than other factors. 
Figure~\ref{fig:feature_weights} (right panel) in Appendix~\ref{app:empirical} shows the distribution of weights given to the five features (race, gender, age, charge degree, and prior counts) when estimating utility ($\vupsilon$). Trends are similar to $\vdelta$ weights, although the spreads are generally larger. Unlike the case of desert, charge degree and prior counts contribute similarly to utility.

\xhdr{Our notion of fairness}
Table~\ref{tab:avg_weights} summarizes the average weights obtained by estimating $\vdelta$ and $\vupsilon$. With circumstance, desert, and utility specified, our EOP-based notion of fairness requires that those who are similarly \emph{deserving}, should have the same prospect for \emph{utility} regardless of their \emph{circumstance} (see Definition~\ref{def:EOP}). Notice that our notion of fairness is much more nuanced than the two definitions of fairness at the center of COMPAS controversy---in particular, many factors beyond the true and predicted labels impact the assessment of fairness in our formulation.  Also observe that while the majority of participants consider race and gender as morally irrelevant in part 1, on average, they believe these factors should play a role in determining a defendant's deserved sentencing decision and the harm/benefit they perceive as a result of their sentencing outcomes.

\xhdr{Variations across gender, race, age, education, \& political views} Tables~\ref{tab:avg_delta_demo}, \ref{tab:avg_upsilon_demo} show the average weights allocated to each attribute of a defendant when $\vdelta$ and $\vupsilon$ are computed for various demographic subgroups of participants. In terms of desert ($\vdelta$), liberal participants give a relatively large positive weight to race, while conservative participants give it a weight close to 0 (indicating that they believe race should not affect whether a defendant deserves leniency). A similar trend holds for female vs. male, non-white vs. white, and young vs. old participants. Younger participants allocate a large positive weight to age (indicating that they believe younger defendants are generally more deserving of leniency), while our older participants allocate a small negative weight to age.

\begin{table}[h]
\caption{Average $\vdelta$ (desert coefficient) for participants belonging to various demographic segments. Significant differences are highlighted.}
\label{tab:avg_delta_demo}
\centering
{ \tiny
\begin{tabular}{l | ccccccc}
Participant  &  Race &  Gender &  Age & Charge & Prior & $y$  \\ \hline \hline
Liberal         & $\textbf{0.16}$ & $-0.22$ &    $0.15$ &        $-0.46$ &       $-0.27$ &        $-0.43$ \\ 
Conservative    & $\textbf{0.01}$ & $-0.15$ &    $0.10$ &        $-0.45$ &       $-0.22$ &        $-0.34$ \\ \hline
High-school     & $0.08$ & $-0.18$ &    $0.11$ &        $-0.51$ &       $-0.31$ &        $-0.40$ \\ 
University      & $0.13$ & $-0.20$ &    $0.14$ &        $-0.44$ &       $-0.22$ &        $-0.41$ \\ \hline
Male            & $\textbf{0.06}$ & $-0.20$ &    $\textbf{0.17}$ &        $-0.42$ &       $-0.27$ &        $-0.41$ \\ 
Female          & $\textbf{0.19}$ & $-0.18$ &    $\textbf{0.07}$ &        $-0.53$ &       $-0.21$ &        $-0.40$ \\ \hline
White           & $\textbf{0.09}$ & $-0.19$ &    $0.11$ &        $-0.48$ &       $-0.28$ &        $-0.45$ \\ 
Non-white       & $\textbf{0.19}$ & $-0.23$ &    $0.20$ &        $-0.41$ &       $-0.17$ &        $-0.29$ \\ \hline
Young (<40)     & $\textbf{0.14}$ & $-0.20$ &    ${\color{red}\textbf{0.18}}$ &        $-0.44$ &       $-0.24$ &        $-0.41$ \\ 
Old             & $\textbf{0.01}$ & $-0.17$ &   ${\color{red}\textbf{-0.03}}$ &        $-0.53$ &       $-0.30$ &        $-0.39$ \\  \hline
\end{tabular}
}
\end{table}

In terms of harm/benefit or utility to defendants, conservative participants believe female defendants benefit more from leniency, while liberal participants give a relatively small (but still negative) weight to gender. Similar to the case of desert, younger participants allocate a large positive weight to age, while older participants give a small negative weight to age.

\begin{table}[h]
\caption{Average $\vupsilon$ (utility coefficient) for participants belonging to various demographic segments. Significant differences are highlighted.}
\label{tab:avg_upsilon_demo}
\centering
{ \tiny
\begin{tabular}{l | ccccccc}
Participant     &  Race &  Gender   &  Age  & Charge    & Prior &  $y$  & $\hy$ \\ \hline \hline
Liberal         &   $0.05$ & $\textbf{-0.06}$ &    $0.08$ &        $-0.10$ &       $-0.08$ &        $\textbf{-0.20}$ &                $-0.47$ \\
Conservative    &   $0.07$ & $\textbf{-0.17}$ &    $0.10$ &        $-0.09$ &       $-0.08$ &        $\textbf{-0.07}$ &                $-0.29$ \\ \hline
High-school     &   $\textbf{0.13}$ & $-0.09$ &    $0.13$ &        $-0.10$ &       $-0.08$ &        $-0.23$ &                $-0.32$ \\
University      &   $\textbf{0.02}$ & $-0.06$ &    $0.06$ &        $-0.09$ &       $-0.08$ &        $-0.15$ &                $-0.48$ \\ \hline
Male            &   $0.06$ & $-0.04$ &    $0.07$ &        $-0.06$ &       $-0.08$ &        $\textbf{-0.13}$ &                $-0.46$ \\
Female          &   $0.05$ & $-0.12$ &    $0.11$ &        $-0.14$ &       $-0.09$ &        $\textbf{-0.26}$ &                $-0.38$ \\ \hline
White           &   $0.06$ & $-0.06$ &    $0.09$ &        $\textbf{-0.05}$ &       $-0.07$ &        $-0.16$ &                $-0.42$ \\
Non-white       &   $0.05$ & $-0.10$ &    $0.07$ &        $\textbf{-0.20}$ &       $-0.11$ &        $-0.23$ &                $-0.45$ \\ \hline
Young (<40)     &   $0.04$ & $-0.09$ &    ${\color{red}\textbf{0.13}}$ &        $\textbf{-0.12}$ &       $-0.08$ &        $-0.18$ &                $-0.43$ \\
Old             &   $0.10$ &  $0.00$ &   ${\color{red}\textbf{-0.06}}$ &         $\textbf{0.00}$ &       $-0.09$ &        $-0.17$ &                $-0.40$ \\ \hline
\end{tabular}
}
\end{table}

\begin{figure}[h!]
    \centering
    \begin{subfigure}[b]{0.23\textwidth}
        \includegraphics[width=\textwidth]{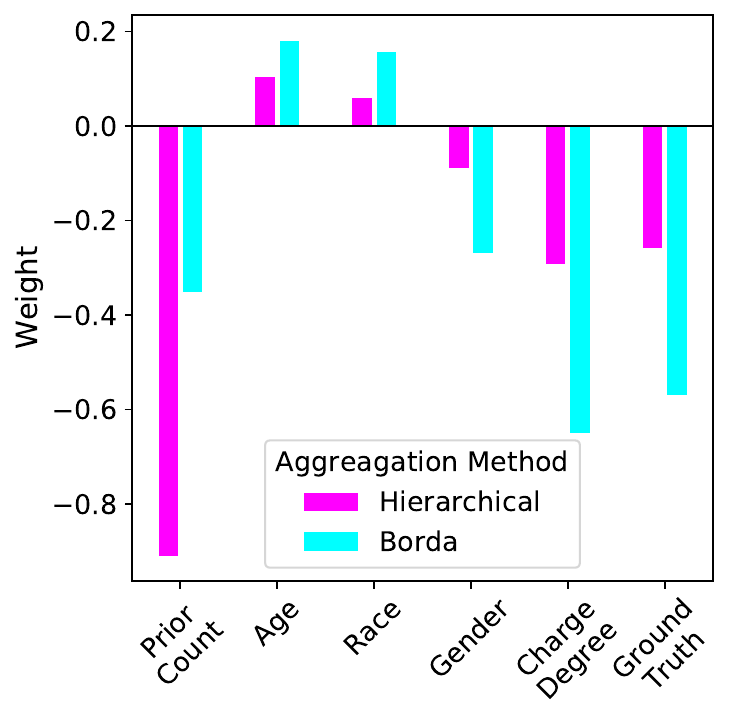}
    \end{subfigure}
        \begin{subfigure}[b]{0.23\textwidth}
        \includegraphics[width=\textwidth]{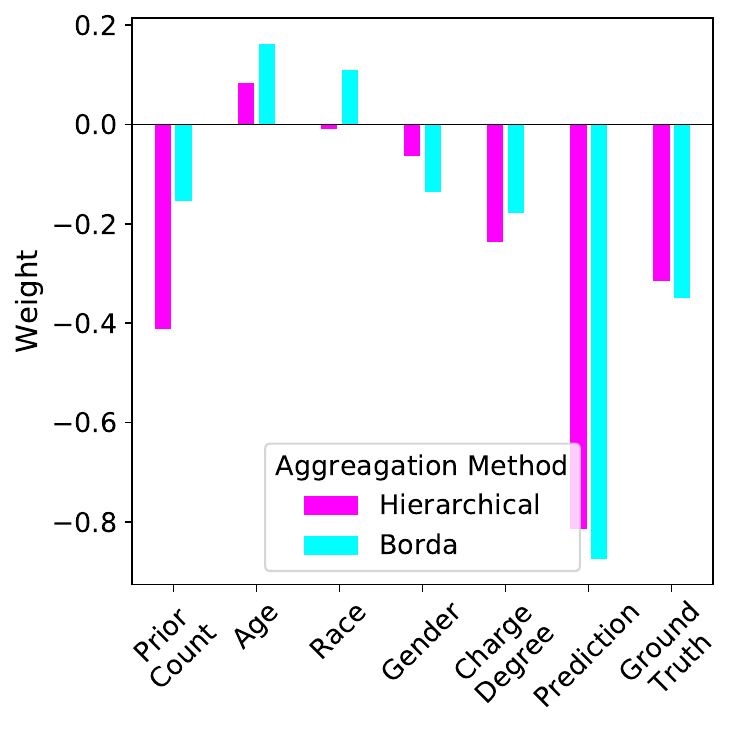}
    \end{subfigure}
    \caption{(Left) $\vdelta$ for the two aggregation methods proposed in Section~\ref{sec:aggregation}. (Right) $\vupsilon$ for the two aggregation methods.}\label{fig:aggregation}
    \vspace{-5mm}
\end{figure}

\xhdr{Sensitivity to the aggregation method}
Figure~\ref{fig:aggregation} contrasts the weights computed according the two aggregation methods described in Section~\ref{sec:aggregation}. While the weights are generally comparable, we observe that the hierarchical model puts a significantly large weight on prior counts. (This is to be expected as this aggregation method zeros in on the points of agreement among participants, and indeed, the majority of participants believe that a defendant with a large prior count is less deserving of leniency.)

\begin{figure}[h!]
    \centering
    \begin{subfigure}[b]{0.23\textwidth}
        \includegraphics[width=\textwidth]{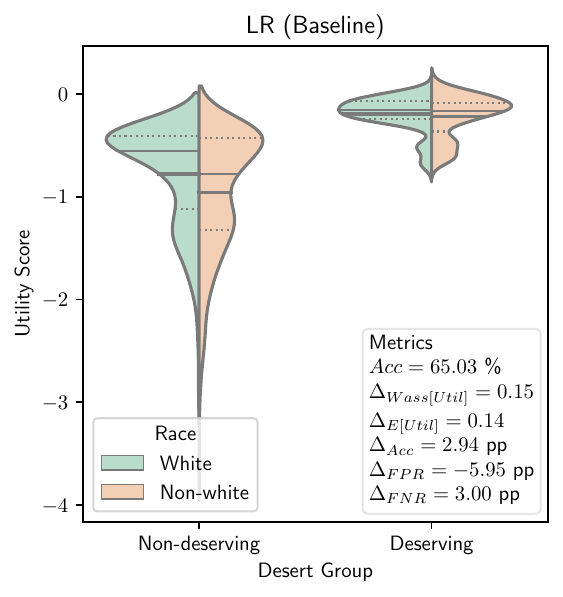}
        \caption{No fairness constraints}
    \end{subfigure}
    \begin{subfigure}[b]{0.23\textwidth}
        \includegraphics[width=\textwidth]{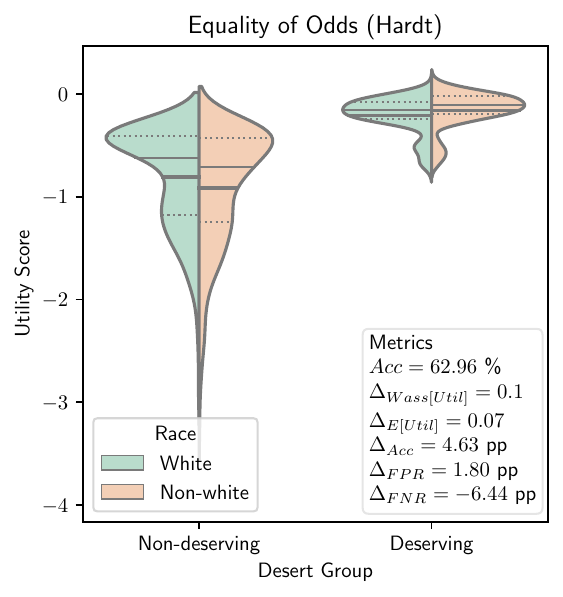}
        \caption{Hardt et al. post-processing}
    \end{subfigure}
    \begin{subfigure}[b]{0.23\textwidth}
        \includegraphics[width=\textwidth]{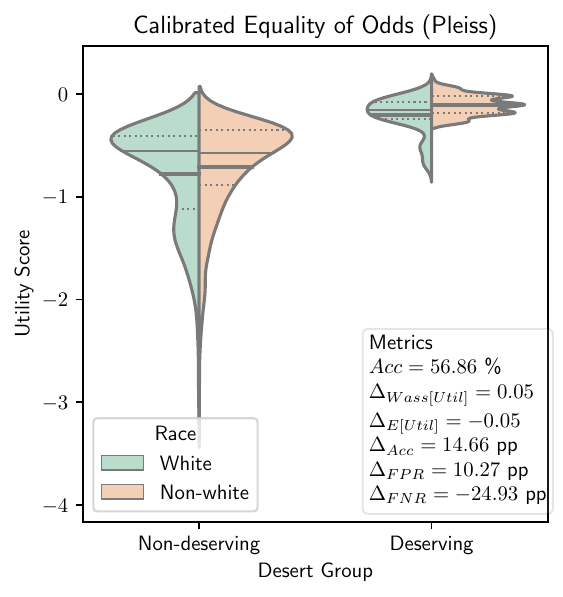}
        \caption{Pleiss et al.'s method}
    \end{subfigure}
    \begin{subfigure}[b]{0.23\textwidth}
        \includegraphics[width=\textwidth]{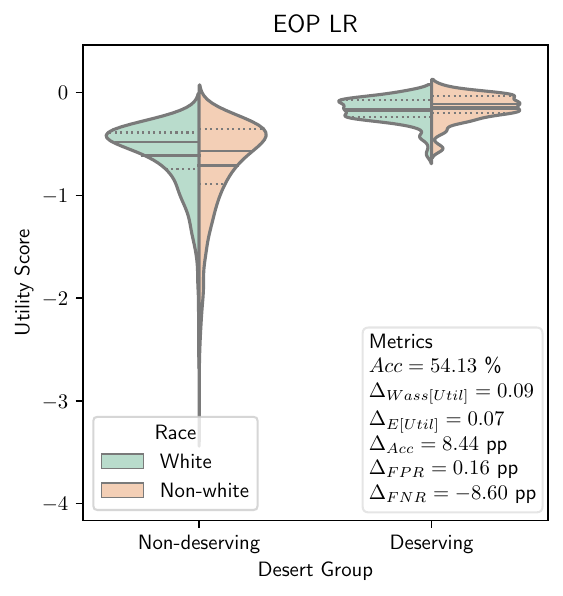}
        \caption{our fairness constraint}
    \end{subfigure} 
    \begin{subfigure}[b]{0.23\textwidth}
        \includegraphics[width=\textwidth]{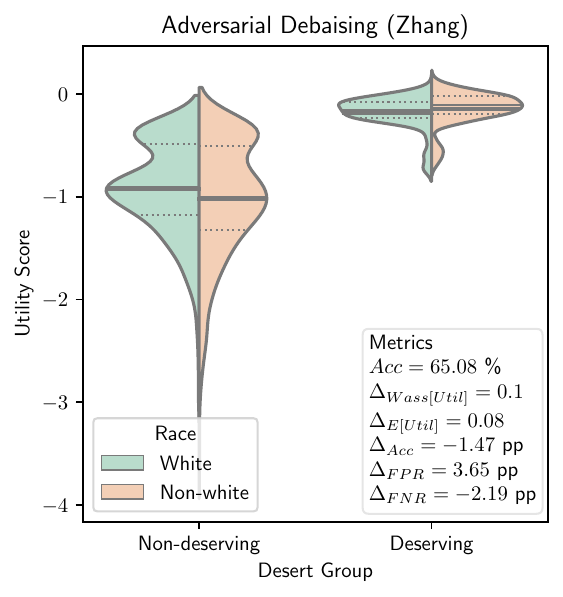}
        \caption{Adversarial debiasing.}
    \end{subfigure}
    \begin{subfigure}[b]{0.23\textwidth}
        \includegraphics[width=\textwidth]{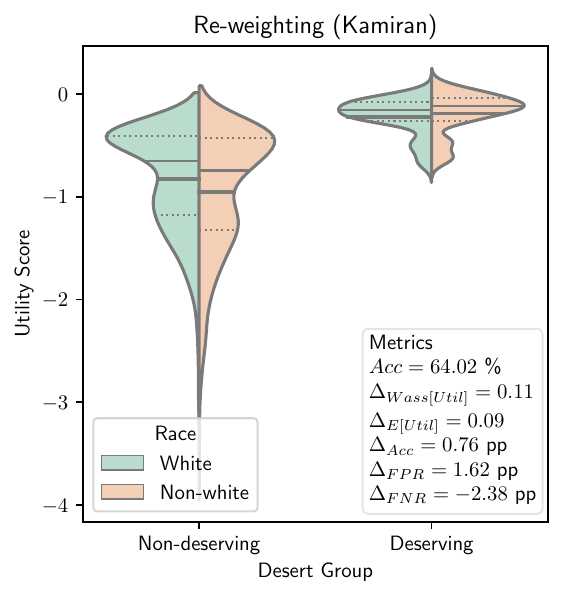}
        \caption{Kamiran et al.'s reweighing}
    \end{subfigure}
    \caption{The distributions of utility across (desert, circumstance) groups when a classifier in trained on the COMPAS dataset using different fairness interventions.}\label{fig:comparison}
    \vspace{-4mm}
\end{figure}

\xhdr{Comparison with existing notions of fairness}
Figure~\ref{fig:comparison} illustrates the effect of existing fairness-enhancing methods on the utility distributions across (desert, circumstance) groups. Note that to simplify the illustration, we have made the desert space binary as follows: a defendant is considered deserving of leniency (or in short, deserving, on the plots) if and only if his/her desert score is above the median score in his/her own circumstance group. This ranked-based approach to desert is compatible with luck egalitarian interpretations of Equality of Opportunity (see e.g.,~\citep{roemer2002equality}).

Figure~\ref{fig:comparison} demonstrates the importance of formulating fairness as the comparison between \emph{distributions}---as opposed to single aggregate measures, such as accuracy, false positive, or false negative rates.
Our key finding here is that compared to other fairness-enhancing methods, our fairness notion performs better in terms of \emph{equalizing the utility distributions} across groups. 
In particular, it pushes up the population to higher utility levels (for example,
75\% of non-deserving and all of deserving subjects receive utilities scores greater than or equal to -1), 
whereas other methods lead to long tails for the non-deserving sub-populations and lower mean utilities overall. For a more detailed comparison, see Appendix~\ref{app:empirical}.

\subsection{Qualitative Findings}
At a high-level, most participants thought race and gender should not be a deciding factor in determining a defendant's sentencing outcome, even if they are statistically correlated with certain types of crimes. Opinions about age were mixed. Some thought younger people are less in control of their decision, so they should be treated with leniency. Others thought a punishment at a younger age is more effective at preventing future crimes. (See Appendix~\ref{app:quote} for the quotations supporting these statements.) 
The majority of participants thought prior convictions should definitely affect a defendant's sentencing outcome. Although, some noted that in a biased judicial system, certain races might have systematically higher numbers of prior convictions. 
In part 2, aside from other features, we provided the participants with the defendants' true labels (i.e., whether the defendant will reoffend in the future). While they generally thought a participant who will not reoffend in the future is more deserving of leniency, some noted that such information is not available at the time of decision-making, so neither defendant should get preferential treatment.
In part 3, we provided the participants with the defendants' predicted labels (i.e., whether the algorithm predicts the defendant has a high risk of reoffending in the future). Participants generally agreed that a low-risk prediction is more beneficial for defendants, while a high-risk prediction is particularly harmful if the defendant does not have a long criminal history and won't reoffend.

Overall, participants found our task thought-provoking and engaging. Our findings indeed showcases their complex reasoning about desert and utility: Our participants took race and gender into account when determining a defendant's desert and utility. For instance, if a defendant is female, participants generally consider her more deserving of a lenient sentencing outcome (compared to an identical male defendant). As another example, participants generally believed that the utility of staying out of jail is higher for a young defendant (compared to an older but otherwise identical defendant). These beliefs are aligned with the participants' perception of the social and economic consequences of sentencing decisions for female and young defendants, respectively (as reflected in their justifications). 

\section{Concluding Discussion}\label{sec:discussion}
Our work addresses a critical gap in the growing literature on Fair-ML: Currently, ML practitioners are the primary designers of mathematical notions of fairness, and the literature lacks a systematic way of bringing human judgment into the process of formulating fairness and translating it into computationally tractable measures.
While our work takes an important step toward empowering people's participation in this process, our simplified framework and small-scale case-study on AMT must be viewed with their limitations and caveats in mind. 
Importantly, any tractable mathematical notion of fairness by definition fails to capture all the nuances and complexities involved in people's ethical reasoning and assessments. \emph{Framing} and \emph{messaging} always play a crucial role in human-subject studies, especially when we attempt to elicit people's moral judgment about a \emph{hypothetical} decision-making scenario. For a detailed discussion of learning-based approaches to moral dilemmas, see~\citep{conitzer2017moral}. We dedicate the rest of this section to elaborate on some of the specific limitations of our work. A more comprehensive discussion (including that of our various simplifying assumptions and modeling choices) can be found in Appendix~\ref{app:limitations}.

\hha{
\xhdr{Whose opinion should matter?}
Attempts to bring human judgment into the process of formulating fairness should not stop at taking input from \emph{ordinary people} (e.g., participants on AMT), for a variety of ethical considerations---chief among them is the principle that a simple majority should not be allowed to determine the rights and freedoms of the minority. In practice, one must strive to involve all stakeholders---including impacted individuals, community members, domain experts, ethicists, policymakers, and system designers---in the process. These parties must collectively deliberate and determine the \emph{weight} of the input from each side (akin to ~\citep{freedman2020adapting}). For example, under some conditions, it may be appropriate to give higher weight to people who have been previously disenfranchised or excluded from deliberations, or those with higher stakes~\citep{woodruff2018qualitative}. There must also exist pathways for overruling certain opinions (e.g., those that violate basic rights, legal requirements, or ethical norms; or those reflecting harmful biases and stereotypes).  Another consequential choice is the mechanism for aggregating opinions. For example, what are the axioms and desiderata that such mechanisms should satisfy~\citep{sen1986social}?  While addressing these vital considerations is out of the current work's scope, they mark critical prerequisites for the ethical deployment of our framework in practice.}

\xhdr{Focus on the EOP family of notions}
We constructed a context-dependent mathematical formulation of outcome fairness by obtaining people's responses to pairwise questions and finding the \emph{EOP-based} fairness notion that best captured those responses. We took it as a given that the EOP family of fairness notions can represent the human judgments of outcome fairness reasonably well---at least in the hypothetical scenario presented to our participants. We acknowledge that real-world scenarios are always highly complex and cannot be faithfully squeezed into any mathematical equation, but that does not justify picking an arbitrary formulation of fairness off the shelf without careful deliberations involving stakeholders. 

\hha{
\xhdr{Causal links among desert, circumstance, and utility} 
We did not directly address the causal relationship between various quantities of interest in our framework. We note, however, that the luck-egalitarian models of equality of opportunity do account for the possibility of circumstance causally impacting desert by defining desert in relevant (quantile-based) terms. For a detailed discussion, we refer the reader to \citep{lefranc2009equality,roemer2002equality,heidari2019a}.
We treated the problem of estimating desert, circumstance, and utility as decoupled problems. A natural direction for future work consists of a holistic treatment of these problems and examining the impact of a joint estimation on our findings.
}

\xhdr{Linear models}
We assumed that both desert and utility functions have a simple linear form.
Linear models simplified our experiments because (1) they are easy to train with a small sample size without the risk of overfitting, and (2) are interpretable. In particular and as shown in Figure~\ref{fig:sims}, we can learn the coefficients of a linear preference vector (up to a constant) by asking only 25 questions on average from each participant.
While the linearity restriction facilitated our illustration,  we must emphasize that in practice, such simplifying choices have ethical implications beyond scalability and interpretability,  so they must be thoroughly analyzed before being used in practice.

\xhdr{Framing Bias}
As with any human-subject experiment, we cannot completely rule out the influence of framing on our empirical observations. We attempted to limit this by ensuring neutral language throughout and reminding the participant they may have a different opinion every time we presented them with a subjective example to demonstrate the task. \hha{We emphasize, though, that framing is not an issue specific to our study--no single human-subject study on its own can claim full-robustness to its experimental setup. Several replications and variation studies of the initial work are always required to establish the robustness of its findings to framing and other psychological effects.}

In conclusion, we believe that the insights offered by our work can inform future studies of this kind, and ultimately, contribute to the proper formulation of fairness in real-world applications of ML. We close by noting that a more comprehensive case-study of our framework must be broader in scope and resources; 
it should carefully provide participants with additional relevant information about decision subjects and inner workings of the decision-making algorithm; and account for the effect of participants' stakes, life experiences, political views, and expertise in their judgments.

\section*{Acknowledgments}
This work was in part done while Heidari was a postdoctoral fellow at ETHZ, and Yaghini was a visiting student from EPFL carrying out his Master's dissertation work at ETHZ.
Heidari acknowledges partial support from NSF IIS2040929. Any opinions, findings, and conclusions, or recommendations expressed in this material are those of the authors and do not necessarily reflect the views of the National Science Foundation and other funding agencies

\pagebreak
\bibliographystyle{ACMReferenceFormat}
\balance
\bibliography{MyBib}

\pagebreak
\appendix

\section{Empirical Details and Findings}\label{app:empirical}

Overall, participants found our task engaging and thought-provoking. For instance, one participant said \textit{``[I] thought it was very interesting and wonder how much an algorithm vs. the human mind really goes into play in possibly life-changing decisions!?!''}
Some noticed that ``in some cases an algorithm might really be unfair'' and hoped that the algorithm bases its consequential predictions on more than 6-7 attributes. 
Several participants mentioned that they enjoyed the conversational format. %
They also appreciated the option to provide us with feedback.

\begin{figure*}[t!]
    \centering
    \includegraphics[width=.9\textwidth]{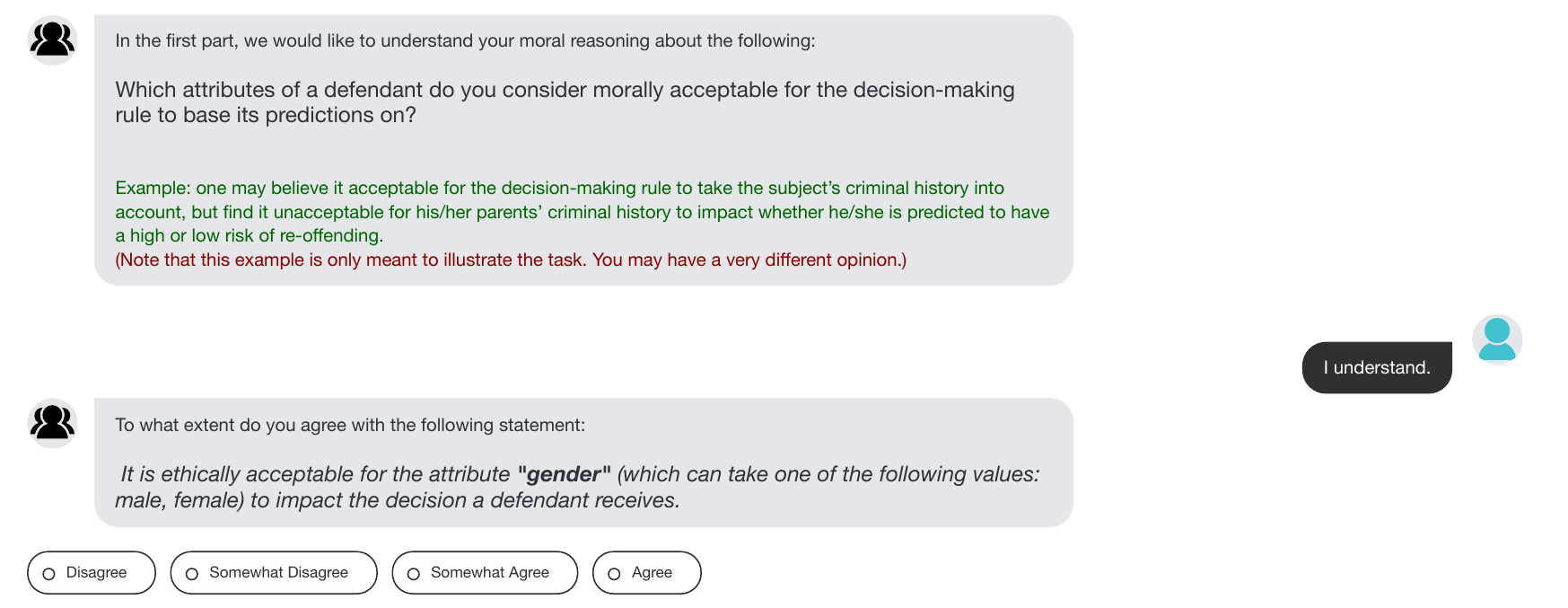}
    \caption{Our conversational user-interface designed to  elicit participants' opinion about desert, utility, and circumstance.}
    \label{fig:UI}
\end{figure*}

\subsection{Justifications}\label{app:quote}

\subsubsection{Response to Part 1}\label{responses1}

\textbf{Race:} Overall, most participants thought race should not be a factor in determining a defendant's sentence, even if it is statistically correlated with certain types of crimes. The following quotes from our participants are meant to show the diversity of opinions regarding race.
\begin{itemize}
\item \textit{``There is no justification for using race as a factor even if it may have statistical significance.''}
\item \textit{``That would sort of be like racially profiling which i don't think is fair.''}
\item \textit{``Using this will only lead to bigger societal problems.''}
\item \textit{``Race shouldn't be a sole deciding factor, but can be used to make a more accurate decisions for re-offenders of certain crimes.''}
\item \textit{``Some races have been discriminated against which has led to higher crime, this would just accentuate that problem.''}
\item \textit{``While majority of people in prison are minorities, I feel it's not fair because it could hurt people who are already disadvantaged.''}
\item \textit{``I think one's economic position does impact sentencing but not race per se.''}
\item \textit{``absolutely, certain groups/races have more of the possibility to re-offend.''}
\end{itemize}

\textbf{Gender:} Overall, most participants thought gender should not be a factor in determining a defendant's sentence, even if it is statistically correlated with certain types of crimes. The following quotes from our participants are meant to show the diversity of opinions regarding gender.
\begin{itemize}
\item \textit{``Males and females should always expect to receive the same version of justice.''}
\item \textit{``Gender is something inherent about a person that cannot be changed. while statistically there are likely differences in commission of crimes by men vs. those committed by women, it is unfair to judge someone on that basis.''}
\item \textit{``Gender is a fluid type of classification, people can even change genders if they wanted to.''}
\item \textit{``Statistically i think that more men offend or re-offend and if that is the case, I think that is acceptable to include this trait in the decision-making rules.''}
\item \textit{``[Gender] is part of the criminal, so it certainly should be considered.''}
\end{itemize}

\textbf{Age:} Opinions about age were mixed. Some thought younger people are less in control of their decision, so they should be treated with leniency. Others thought a punishment at a younger age is more effective at preventing future crimes. The following quotes from our participants are meant to show the diversity of opinions regarding gender.
\begin{itemize}
\item \textit{``The criteria should be about cognitive ability, not age, as such. young people are believed less responsible for decision-making and the justice system is and should consider that in decisions.''}
\item \textit{``If they are younger it means their brain isn't fully developed and you should consider that when evaluating their decision making process.''}
\item \textit{``Minor crimes should have a lesser charge but major crimes should not take age into account. [Leniency toward the youth] carries a risk as well of younger people thinking they can do what they want with less of a punishment when in fact they age at an age where the punishment is more likely to teach them a lesson.''}
\item \textit{``Older people are probably less likely to reoffend.''}
\item \textit{``An older suspect would, on average, live less years, so would have less chance to reoffend.''}
\item \textit{``Good to make an impact at an early age so one does not reoffend.''}
\end{itemize}

\textbf{Prior Counts:} The majority of participants thought prior convictions should clearly affect a defendant's sentencing. Although, some noted that in a biased judicial system, certain races may have systematically higher numbers of prior convictions. The following quotes from our participants are meant to show the diversity of opinions regarding prior counts.
\begin{itemize}
\item \textit{``If someone can commit harsh crimes, they probably don't care about others or themselves and will re-offend.''}
\item \textit{``If one has been convicted before but continues breaking the law, then it is logical to assume this will continue.''}
\item \textit{``Previous offenses is directly about that specific person's history and not a general data set based on a group (age, sex, race, etc), so I believe it's fair to judge someone based on their own previous history.''}
\item \textit{``Prior counts are a result of the racially biased criminal justice system. If something could be done to counterbalance inherent racism in the training data, then prior counts could be useful and ethically acceptable.''}
\end{itemize}

\textbf{Charge Degree:} Similar to prior counts, most participants thought charge degree must be an important factor in sentencing decisions.
\begin{itemize}
\item \textit{``A serious crime which involves violence especially requires a longer and more severe restriction on the defendant.''}
\item \textit{``If the racially biased training data could be avoided, then charge degree would be ethically acceptable to use in the system.''}
\end{itemize} 

\subsubsection{Response to Part 2}\label{responses2}
The justifications provided based on the defendant's attributes, such as race, age, and prior counts, generally reflect the same trend as mentioned in Section~\ref{responses1}. See the following quotes, as examples:
\begin{itemize}
\item \textit{``There wasn't much difference between the two subjects so I think that the non-white person would slightly get more benefit.''}
\item \textit{``The way our court systems are set up most whites benefit a lot more than non-whites.''}
\item \textit{``This subject is younger with less prior counts so will very much likely benefit from a better life.''}
\item \textit{``Assuming gender worked against subject 1, he caught a break.''}

\end{itemize}

In addition to these attributes, in part 2 we provided the participants with the true labels (i.e., whether each defendant will reoffend in the future). While they generally thought a participant who will not reoffend in the future is more deserving of leniency, some noted that such information is not available at the time of decision-making, so neither defendant should get a preferential treatment.

The following quotes from our participants are meant to show the diversity of opinions regarding the role of true outcome in determining deserved sentencing decision.
\begin{itemize}
\item \textit{``With the benefit of knowing the actual outcome, it is easy to say that subject 2 deserves the more lenient decision.''}
\item \textit{``You can be more lenient if they won't offend again.''}
\item \textit{``Will not reoffend, so in theory, the system should be more likely to give them a more lenient decision.''}
\item \textit{``Ultimately i believe the severity of the crime is the most important factor for the punishment besides that whether or not the reoffend would be second, but that is not possible to know ahead of time.''}
\item \textit{``Hindsight is 20/20 so I had to choose the person which did not re-offend, however at the time of sentencing these two [defendants] would be utterly even.''}
\item \textit{``since [subject] 1 is not white, I want to be lenient due to how minorities are put in prison at such large rates, but I had to consider if they will reoffend.''}
\item \textit{``You don't have actual outcome making the decision, not sure if it belongs in the question here.''}
\item \textit{``I'm ignoring the actual outcome because they both have the same data.''}
\end{itemize} 

\subsubsection{Response to Part 3}
The justifications provided based on the defendant's attributes, such as race, age, and prior counts, generally reflect the same trend as mentioned in Section~\ref{responses1}, indicating that according to our participants, historically disadvantaged groups generally benefit more from lenient decisions and are harmed more by harsh sentencing decisions. In addition to these attributes, in part 3 we provided the participants with the true labels and predicted labels (i.e. whether the algorithm predicts each defendant has a high risk of reoffending in the future). Pariticpants generally agreed that a low-risk prediction is more beneficial to the defendant, while a high-risk prediction is particularly harmful to the defendant if he/she does not have a long criminal history and is not going to reoffend.

The following quotes from our participants are meant to show the diversity of opinions regarding the role of the predicted label in determining deserved sentencing decision.
\begin{itemize}
\item \textit{``Subject 2 was considered low risk to reoffend, so they would probably get a lesser sentence or no jail time. So they would benefit a lot more.''}
\item \textit{``They were listed as low risk and were in fact low risk, so it helped them out a lot.''}
\item \textit{``even with a felony charge, this subject still got a low risk from the algorithm so I think he benefited more from it's decision.''}
\item \textit{``[Subject] 2 would benefit more in being given the benefit of the doubt when it was not actually warranted.''}
\item \textit{``because subject 1 committed a felony and the algo concluded that subject 1 was at a low risk to reoffend. meanwhile subject 2 only committed a misdemeanor with all other attributes the same, yet the algo concluded that subject 2 was at a high risk to reoffend. so clearly subject 1 would benefit more from the algo's decision.''}
\item \textit{``they are listed as low risk but will reoffend so it helped them personally as a criminal to be considered low risk.''}
\item \textit{``if the algorithm says someone has a low risk to offend and they actually do reoffend, then putting back on the streets gives them an opportunity to do what they want to do.''}
\item \textit{``subject 1 would benefit less because he has a lesser charge degree, but still gets the decision "high risk to reoffend."''}
\end{itemize}

\begin{figure}[t!]
	\centering
	\begin{subfigure}[b]{0.23\textwidth}
		\includegraphics[width=\textwidth]{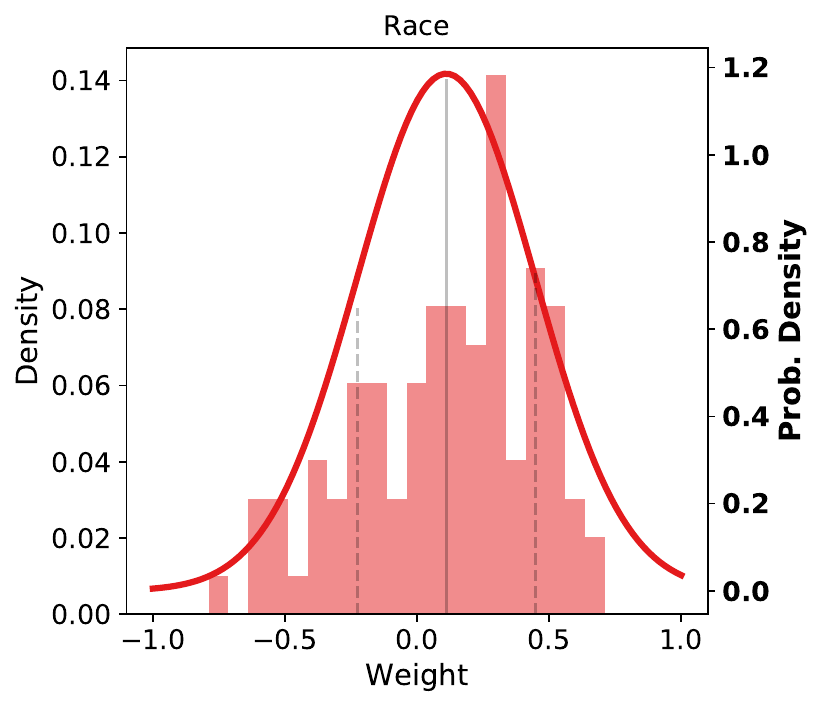}
	\end{subfigure}
		\begin{subfigure}[b]{0.23\textwidth}
		\includegraphics[width=\textwidth]{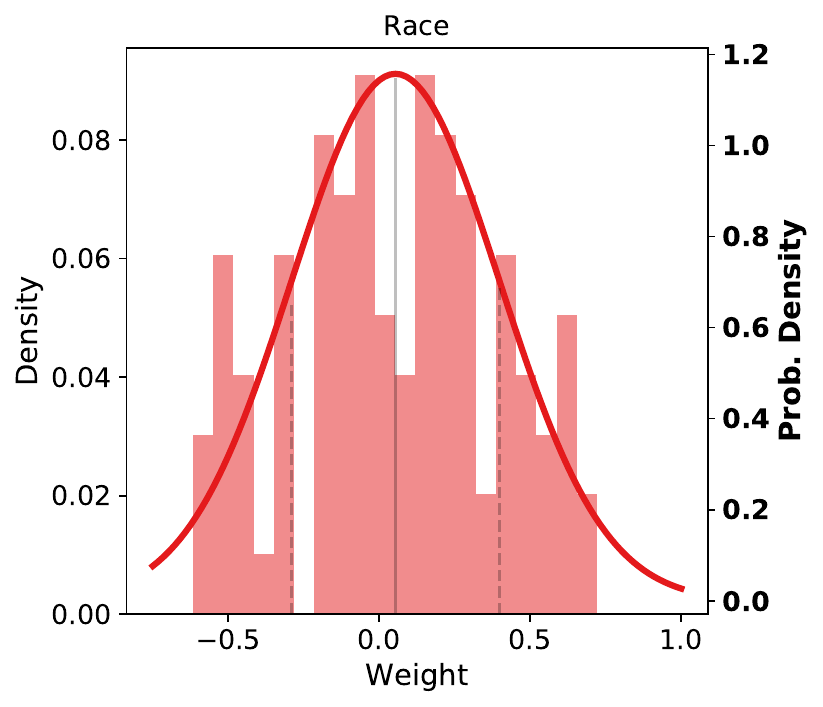}
	\end{subfigure}
	\begin{subfigure}[b]{0.23\textwidth}
		\includegraphics[width=\textwidth]{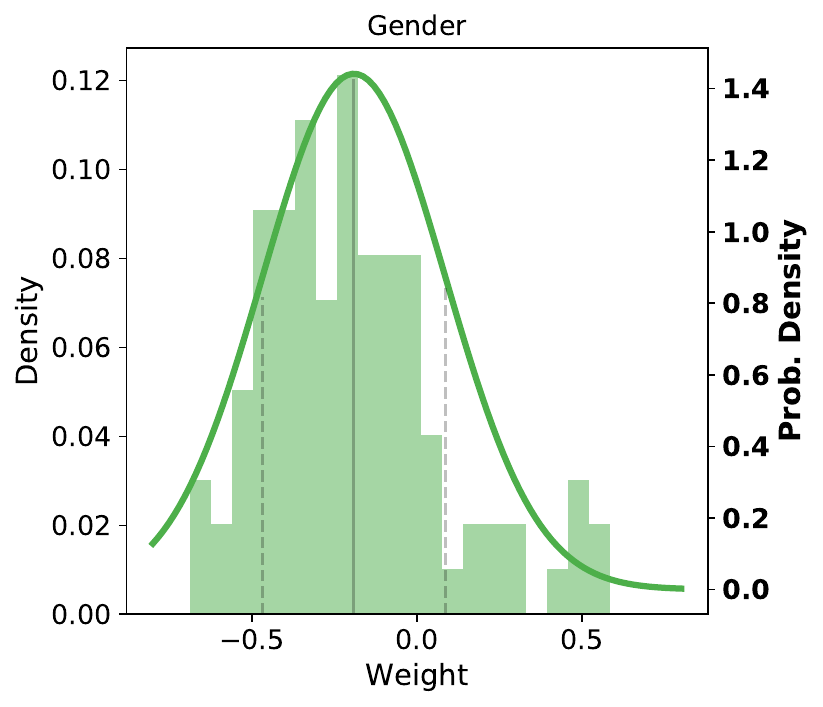}
	\end{subfigure}
		\begin{subfigure}[b]{0.23\textwidth}
		\includegraphics[width=\textwidth]{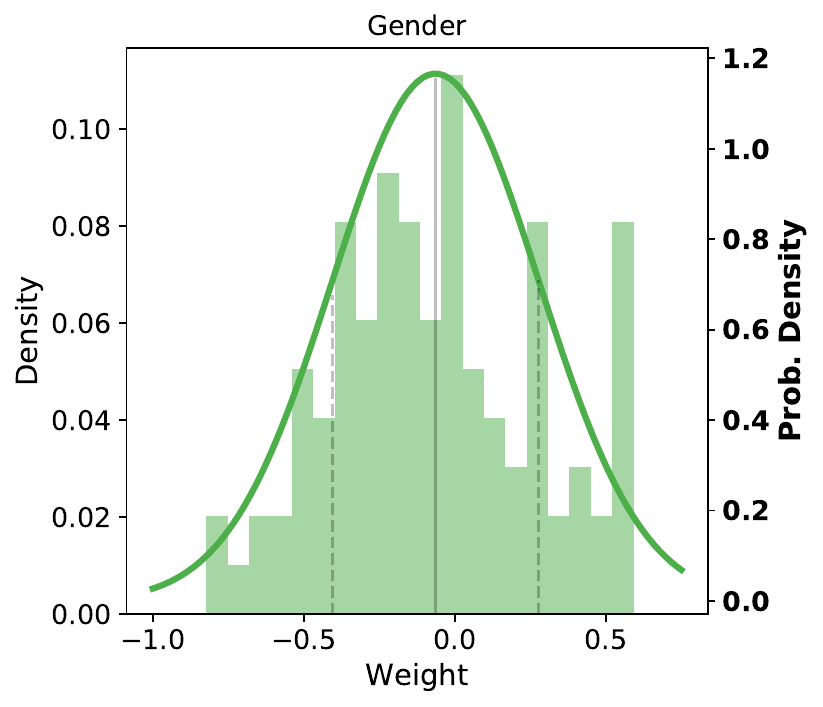}
	\end{subfigure}
	\begin{subfigure}[b]{0.23\textwidth}
		\includegraphics[width=\textwidth]{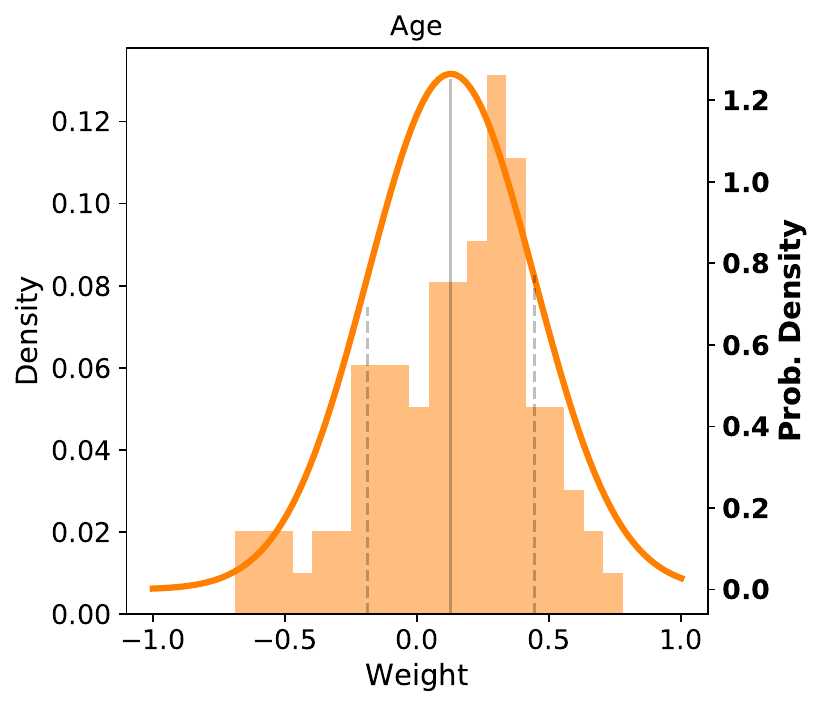}
	\end{subfigure}
		\begin{subfigure}[b]{0.23\textwidth}
		\includegraphics[width=\textwidth]{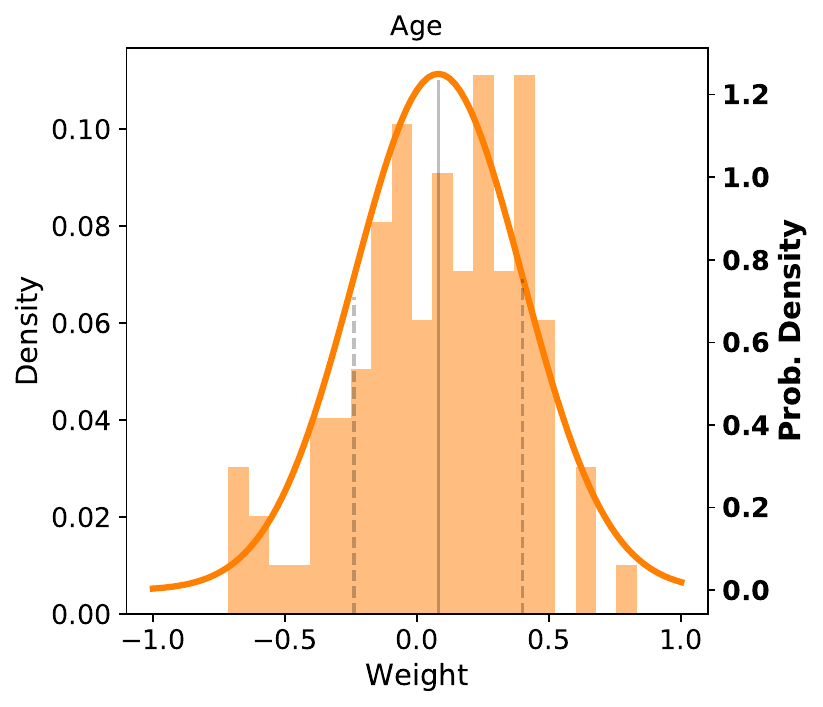}
	\end{subfigure}
	\begin{subfigure}[b]{0.23\textwidth}
		\includegraphics[width=\textwidth]{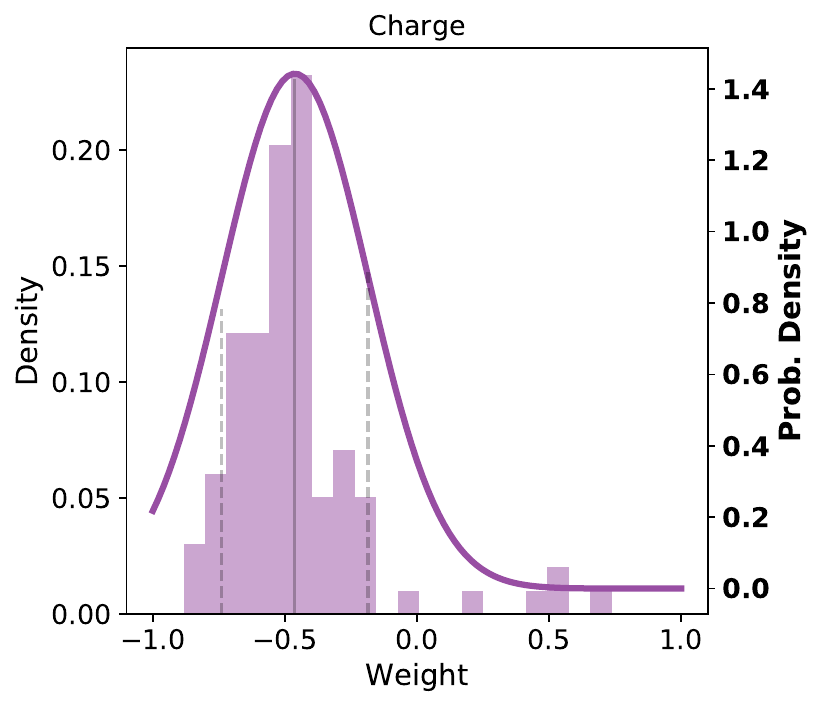}
	\end{subfigure}		
	\begin{subfigure}[b]{0.23\textwidth}
		\includegraphics[width=\textwidth]{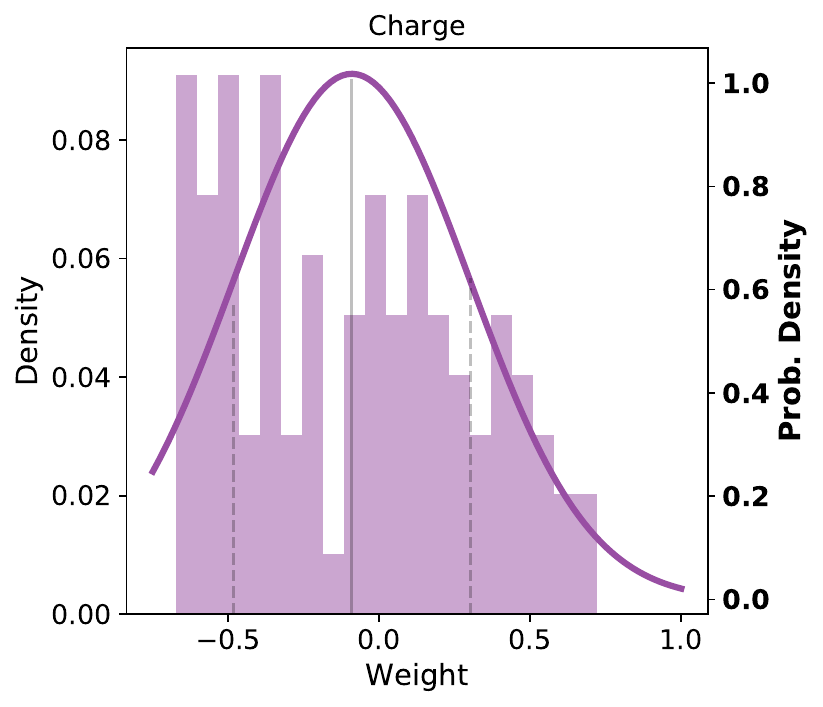}
	\end{subfigure}		
	\begin{subfigure}[b]{0.23\textwidth}
		\includegraphics[width=\textwidth]{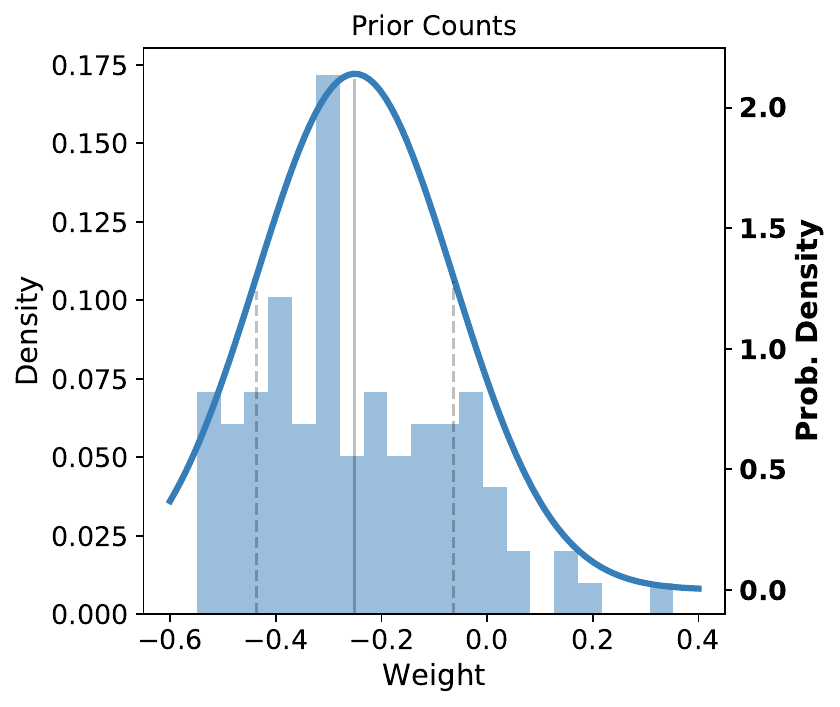}
	\end{subfigure}	
	\begin{subfigure}[b]{0.23\textwidth}
		\includegraphics[width=\textwidth]{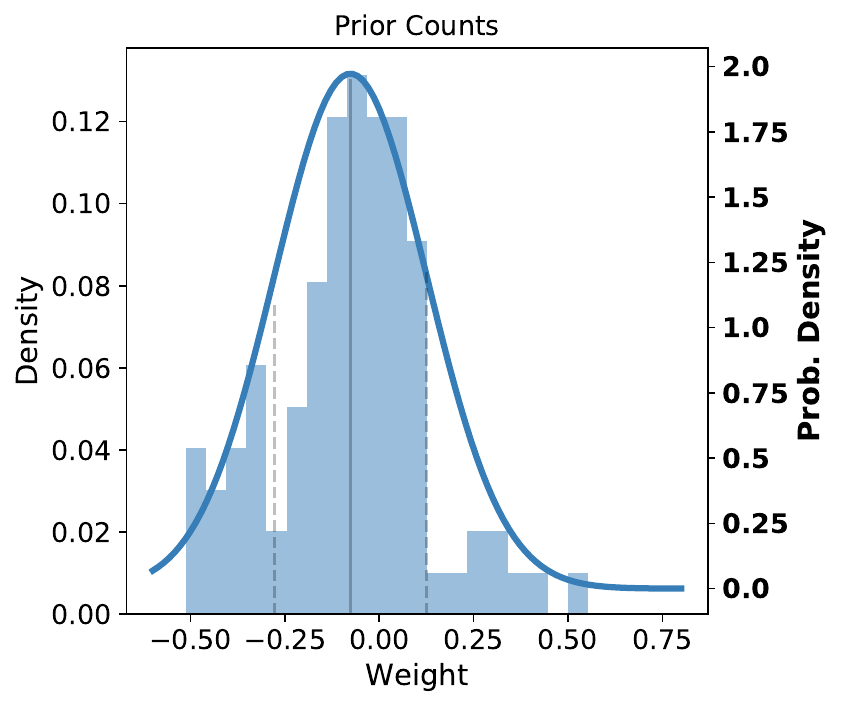}
	\end{subfigure}	
	\caption{(Left) From top to bottom: the empirical distribution of $\delta$ on various features when $\vdelta$ is estimated for each of our 99 participants, separately. (Right) The distribution of $\upsilon$ for the same setup.
	}\label{fig:feature_weights}
\end{figure}

\section{Limitations}\label{app:limitations}

\xhdr{Engagement on AMT}
One barrier to obtaining meaningful answers from participants is to maintain participants' engagement and attention to the task. 
To prevent the possibility of participants choosing their answers without considering the information provided to them---and in the worst case, completely at random, we took the following steps:
(1) Restricting participation to Turkers with $99\%$ approval rate.
(2) Limiting the number of questions to the necessary minimum.
(3) Incentivizing the participant to justify their answers.
(4) Adding randomly-placed attention-check questions to our questionnaires.
(5) Restricting participants to complete the task at most once.

\hhl{
\xhdr{Limited number of features}
We selected the specific set of observable features from the COMPAS data set for the following two reasons. First, participants could easily understand these features. Second, this small subset of attributes would cover a wide range of moral and social relevance. We expected some of them to be considered arbitrary by the majority of our participants (e.g., race), some highly relevant (e.g., prior counts), and some in between (e.g., age). In practical situations, there must be a thorough deliberation on what features should contribute to each EOP-parameter and in what form.
}

\hhl{
\xhdr{The standard Gaussian CDF for noise}
Note that when modeling human behavior, we cannot  expect the model to predict the behavior of the human subject, deterministically. Instead, the model must be able to accommodate noisy observations. We utilized a standard Gaussian/normal distribution for noise because it is a common choice in modeling individual pair-wise choices and has been previously adopted in the literature. In practice, if domain-knowledge prescribes a specific distribution for noise, that distribution should replace the normal CDF used in our illustration.
}

\hhl{
\xhdr{Our use of the COMPAS dataset} 
Our use of the COMPAS data set was solely for the purpose of illustrating our framework on a dataset grounded in the real-world. We made several simplifying assumptions about the hypothetical decision-making context and the dataset. For instance, we assumed the dataset is unbiased. Dealing with the biases encoded in this specific dataset---while highly important---falls out of the scope of this work.
As another example, we provided a crude definition for features, such as prior counts, to our participants . Importantly, we did \emph{not} distinguish between different types of prior convictions---in part to keep the survey simple and in part because we did not have access to such fine-grained data. 
}

\hhl{
\xhdr{Biased training data}
Throughout, we focus on a simplified setting in which we assume the true label is unbiased. In real-world settings (e.g., in case of COMPAS), the training data itself may be biased. While our framework would allow participants to partially address such biases (e.g., by adjusting the coefficient of race in the desert function to partially combat the existing biases against certain races in the judicial system), addressing the biases in the data is in itself a substantial challenge that falls out of the scope of this work. 
}

\hhl{
\xhdr{The absence/presence of a neutral response} 
We intentionally did not provide a neutral response option to our participants to block them from taking a neutral moral stance. Instead, we provided them with the option to verbalize their opinion through free-form text boxes after each question. 
We decided against providing a neutral option in our main experiment for two reasons. First, we worried that it would dissuade the participants from an in-depth analysis of the dilemma put in front of them (i.e., comparing the two defendants in terms of desert or utility). Second, and on the technical side, a neutral response would not provide the model with any additional information, forcing us to increase the number of questions. We repeated our experiments (at a smaller scale) including a neutral response option in the user-interface, and did not observe significant differences in the estimated weight vectors. Our findings can be found in Appendix~\ref{app:additional}. 
}

\hhl{
\xhdr{On the incomparability of defendants}
Note that throughout our experiments, participants had the option to provide a written free-form response to questions. The number of cases in which they indicated that they could not make the comparison was negligible.
}

\hhl{
\xhdr{Small sample size} 
While we  provide a separate analysis of the data gathered from different demographic groups, we emphasize that because of the small sample size, the obtained results do not necessarily generalize to the entire population and are for illustration purposes only. 
}

\xhdr{Hypothetical context}
Last but not least, we focused on a simplified hypothetical context to illustrate our framework. We hope that further experiments in other real-world decision-making contexts will further establish the power of our framework in constructing a context-dependent formulation of fairness.

\section{Additional Experiments}\label{app:additional}
\begin{figure}[h!]
    \centering
    \includegraphics[width=.3\textwidth]{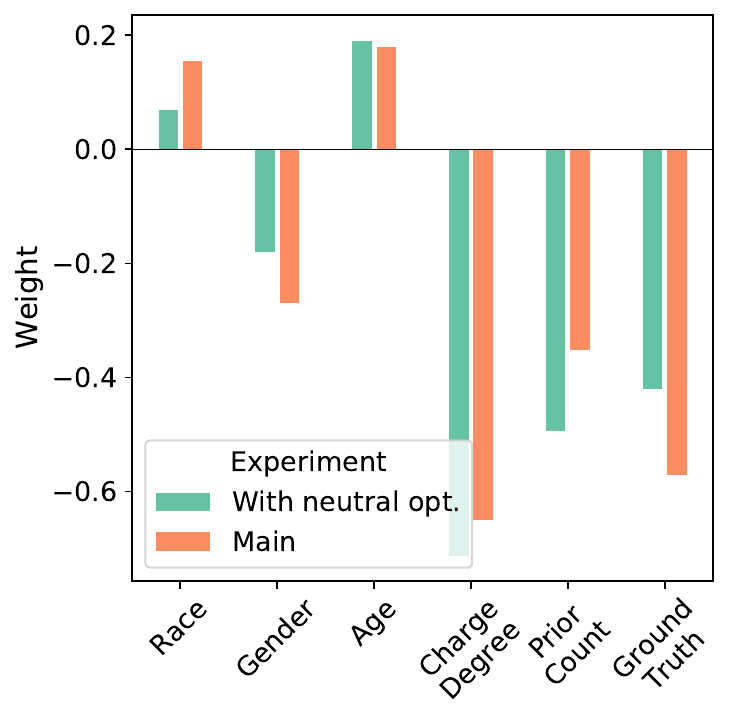}
    \caption{The weight vector, $\vdelta$, estimated using the responses of 20 AMT participants through a modified UI with a neutral response option.}
    \label{fig:neutral}
\end{figure}

We repeated our experiments with a modified UI, in which we provided a neutral response option to participants. Figure~\ref{fig:neutral} shows the weight vectors obtained through this UI using the responses of 20 AMT participants. We do not observe significant differences between the weight vectors learned through our original UI and the modified version. 

\begin{figure}[h!]
    \centering
    \includegraphics[width=.3\textwidth]{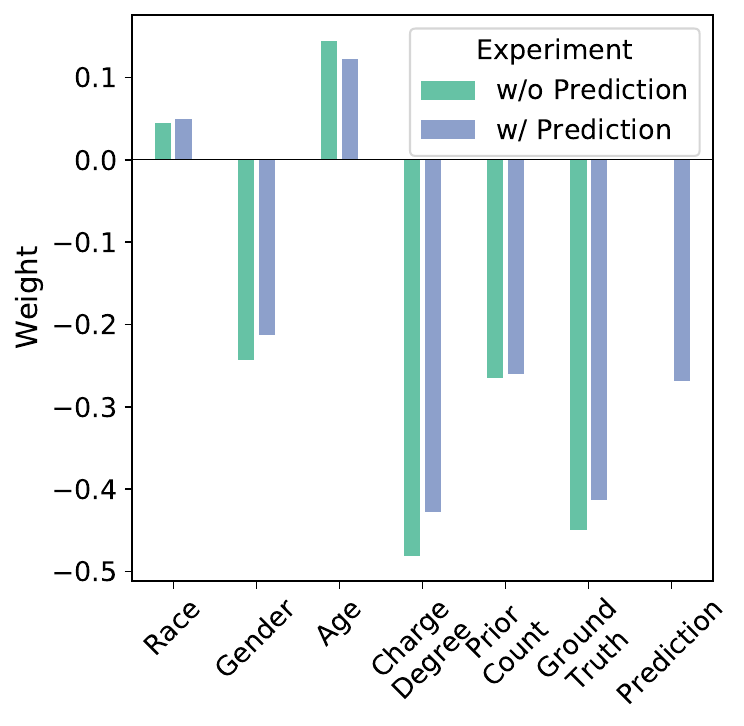}
    \caption{The weight vector, $\vdelta$, estimated using the responses of 20 AMT participants---provided to a modified version of our second questionnaire, in which participants can observe defendants' predicted label. The log-likelihood of observing this data is -7.17. If we exclude $\hy$, the likelihood reduces to -8.00.}
    \label{fig:yhat}
\end{figure}

We also repeated our experiments with a modified version of our second questionnaire, in which participants were given access to defendants' predicted label (as well as their attributes and true labels). The weight vectors obtained using the responses of 20 AMT participants are displayed in Figure~\ref{fig:yhat}. We do not observe significant differences between the weight vectors learned through our original UI and the modified version.

\begin{figure*}[h!]
  \centering
  \includegraphics[width=\textwidth]{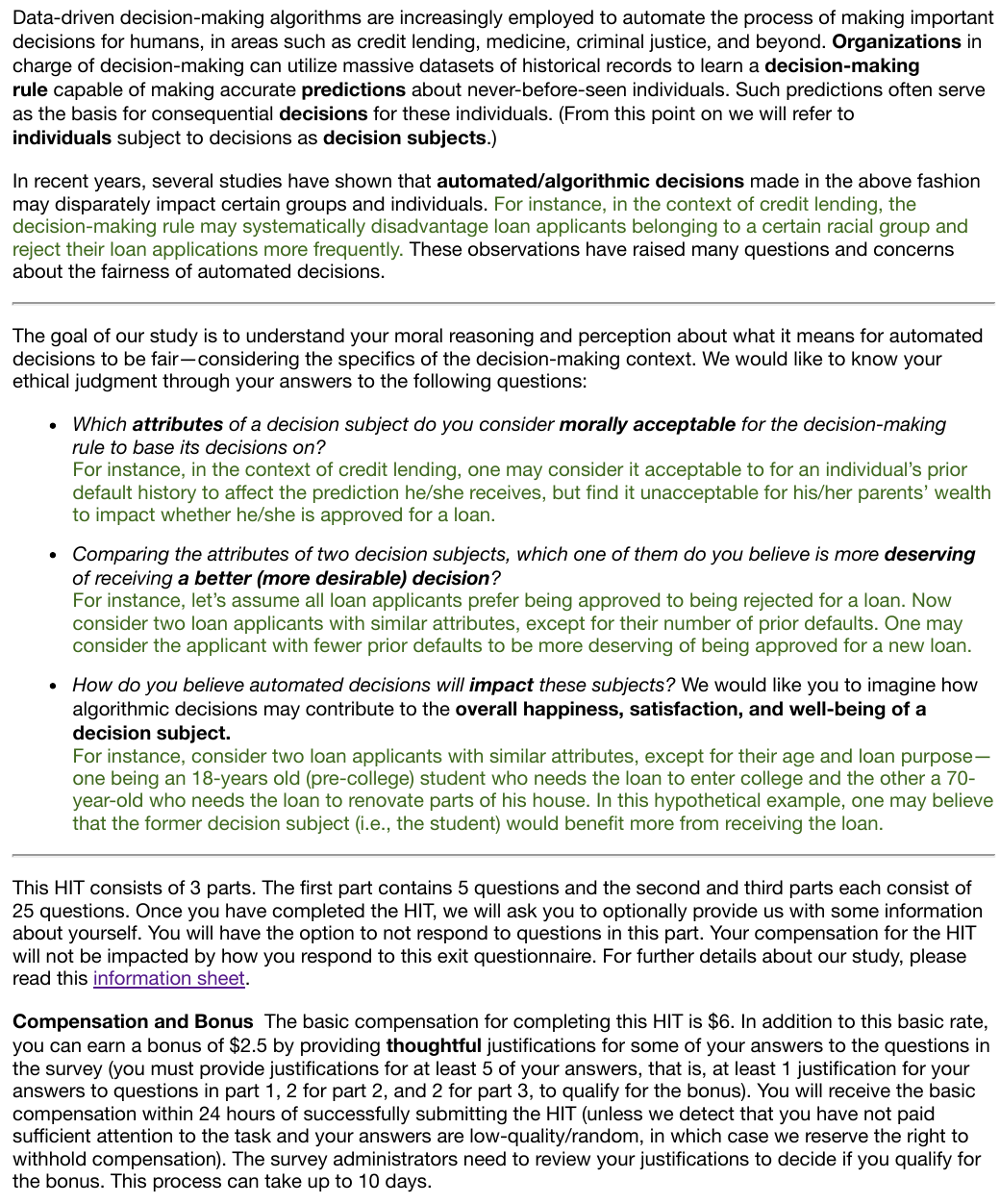}
  \caption{A more detailed version of the introduction to the HIT.}
  \label{fig:long_intro}
\end{figure*}

\end{document}